\documentclass[11pt]{article}

\usepackage[final]{acl}

\usepackage{times}
\usepackage{latexsym}
\usepackage{amsmath}
\usepackage{array}
\usepackage{tabularx}
\usepackage{multirow}
\usepackage{booktabs}
\usepackage{colortbl}

\usepackage[T1]{fontenc}
\usepackage[utf8]{inputenc}

\usepackage{microtype}

\usepackage{inconsolata}

\usepackage{graphicx}
\usepackage{float}

\title{MM-SystemOneBias: Evaluating Bias in Multimodal Jev-Style System One Decision Models}

\author{Jinzhe Li$^{1,3}$ \quad 
        Qi Wang$^{1}$ \quad
        Dandan Guo$^{1}$ \quad 
        Li Liu$^{4}$ \quad  
        Yuan Wu$^{1}$\thanks{Corresponding author} \quad
        Yi Chang$^{1,2,3}$ \\
        $^{1}$School of Artificial Intelligence, Jilin University \\
        $^{2}$Engineering Research Center of Knowledge-Driven Human-Machine Intelligence, Jilin University \\
        $^{3}$International Center of Future Science, Jilin University \\
        $^{4}$School of Big Data and Software Engineering, Chongqing University \\
         \{jinzhe25\}@mails.jlu.edu.cn, dcsliuli@cqu.edu.cn, \{qiwang, guodandan, yuanwu, yichang\}@jlu.edu.cn \\
  }

\begin{document}
\maketitle
\begin{abstract}
Reliable multimodal decision models must preserve semantic decisions across presentation changes, ignore irrelevant additions, and distinguish supporting from conflicting text, yet existing studies examine decision-interface biases and multimodal behavior separately without matched backbone comparisons. We introduce \textbf{MM-SystemOneBias}, evaluating 8 decision models and 5 backbones on 5,323 original instances across 9 datasets, 11 task scenarios, and 6 interventions: option order, option label, external judgment, image--text description, irrelevant information, and rating scale. Decision models remain sensitive across interventions: incorrect external judgments introduce errors in 20.14\% of originally correct responses on average, conflicting descriptions redirect decisions even when rejected in isolation, and answer probabilities on unchanged decisions show mean absolute shifts of 2.92\% to 8.97\%. Relative to backbones, decision models improve presentation, rating-scale, and irrelevant-information stability, yet face a trade-off on added text: they lose less accuracy to misleading cues while gaining less from correct judgments and often losing accuracy under consistent descriptions. These findings highlight visual grounding and probability robustness as key priorities.The code and dataset is available at \url{https://github.com/ALIENS32/MM-SystemOneBias}.
\end{abstract}

\begin{figure}[!t]
    \centering
    \includegraphics[width=0.8\columnwidth]{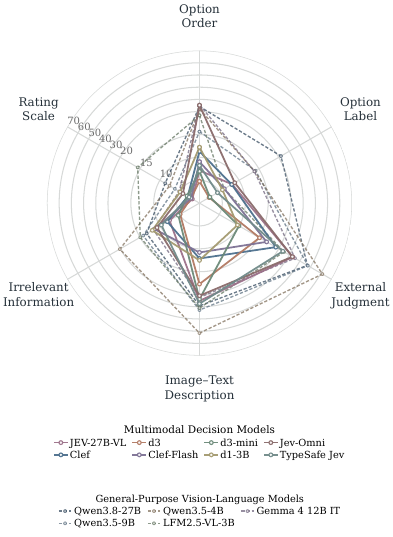}
    \caption{\textbf{Intervention sensitivity across decision models and general-purpose backbones.} Axes show semantic flip rate (option order, option label, irrelevant information), error introduction rate (incorrect external judgments, conflicting descriptions), and normalized grade shift (rating scale; \%, piecewise scale, lower is better).}
    \label{fig:intro_bias_radar}
\end{figure}

\section{Introduction}

Recent Jev-style System One decision models make fast, typed decisions over finite answer spaces practical for structured tasks~\citep{deusser2026evaluating,geuter2026innerjev,model_typesafe_jev}. Rather than generating free-form text or reasoning chains, they map an input directly onto a constrained semantic answer space (a choice, binary judgment, or ordered score) and natively output a probability distribution over legal answers. Multimodal System One decision models extend this interface to applications such as web interaction and image assessment~\citep{data_mind2web,data_agiqa,yue2024mmmu,fu2023mme}, where decisions directly determine actions, moderation outcomes, or quality grades; reliability therefore matters beyond accuracy on a fixed presentation.

A reliable multimodal decision model should preserve its semantic answer across presentation changes, ignore irrelevant additions, and respond appropriately to added textual claims. This requirement is distinct from general multimodal capability: a model may recognize the visual content yet still alter its decision under an intervention.

Existing evaluations leave this behavior undercharacterized. Studies of typed and multiple-choice decision models examine position, label, and scale biases~\citep{sun2026typesafeerrorfree,azizi2026labelsoverride,gao2026choicesfewerdecisions,zheng2023robustselectors,zhao2021calibrate,lu2022fantastically,liusie2023worldknowledge}, while multimodal bias studies primarily evaluate general-purpose vision-language models or MLLM-as-a-Judge systems under open-ended generation, question answering, or scoring~\citep{lee-etal-2026-mm,zhou2026moground,lai2026multimodalseebefore,li2023pope,chen2024unimodalbias,cai2025modalityinterference}. These settings differ from finite semantic decisions on practical multimodal tasks and lack a common protocol comparing decision models with their corresponding backbones.

We introduce \textbf{MM-SystemOneBias}, a benchmark for bias in multimodal decision models spanning 5,323 original instances across 9 public datasets and 11 task scenarios. It evaluates 8 decision models (7 open-weight, 1 closed-source) and 5 general-purpose vision-language models across 6 controlled intervention types: option order, option label, external judgment, image--text description, irrelevant information, and rating scale. The benchmark jointly tracks semantic decisions, task accuracy, and answer probabilities across 7 matched comparisons over 5 backbones.

Decision models show heterogeneous sensitivity across interventions (Figure~\ref{fig:intro_bias_radar}). Relative to their backbones, they are uniformly more stable under option-label, irrelevant-information, and rating-scale interventions, and flip less under option reordering in 6 of 7 comparisons. On added text, however, they face a trade-off: they lose less accuracy to incorrect judgments and conflicting descriptions, yet gain less from correct judgments and lose net accuracy on average under consistent descriptions. Conflicting descriptions still induce task errors even when rejected in an independent visual verification probe, though an image-first instruction mitigates this gap. Even without flipping decisions, interventions shift answer probabilities in decision models by 2.92\% to 8.97\% on average.

Our contributions are:
\begin{itemize}
    \item We introduce \textbf{MM-SystemOneBias}, evaluating 6 controlled presentation and added-text interventions in multimodal decision models.
    \item We conduct matched comparisons between multimodal decision models and their corresponding general-purpose backbones across all 6 intervention types.
    \item We jointly analyze semantic decisions, task outcomes, and answer probabilities to capture both decision flips and probability shifts within stable answers.
\end{itemize}

\section{Related Work}
\label{sec:related_work}

\subsection{Bias in System One Decision Models}

System One decision models return a decision over a typed, finite answer space, often alongside legal-answer probabilities. Prior work establishes capability and probability-quality baselines on structured tasks~\citep{deusser2026evaluating,geuter2026innerjev,guo2017calibration,kadavath2022know}, while interface-bias studies show that name--definition swaps, answer-label changes, and rating-scale shifts can alter equivalent decisions~\citep{sun2026typesafeerrorfree,azizi2026labelsoverride,gao2026choicesfewerdecisions,zhao2021calibrate,lu2022fantastically,liusie2023worldknowledge}. Because these studies focus mainly on text or single intervention families, we extend them to multimodal tasks across 6 controlled intervention types.

\subsection{Bias in General-Purpose Language and Vision-Language Models}

Parallel work studies bias and evidence use in general-purpose language and vision-language models~\citep{liu2023llava,fu2023mme,yue2024mmmu,li2024mllmsurvey}. Controlled interventions document sensitivity to option position and labels~\citep{zheng2023robustselectors,atabuzzaman2025selectionbias,zhao2021calibrate,lu2022fantastically}, as well as irrelevant text, conflicting image--text evidence, and rating formats~\citep{zhou2026moground,wang2026irrelevanttext,lai2026multimodalseebefore,jo2026changenumber,li2023pope,chen2024unimodalbias,cai2025modalityinterference}. Multimodal-judge benchmarks likewise test how input perturbations shift scores and preferences~\citep{ye2024justice,hwang2025fooling,lee-etal-2026-mm,zheng2023mtbench,shi2024judgingjudges,wataoka2024selfpreference}. Unlike open-ended generation or judge settings, we evaluate multimodal decision models alongside their corresponding backbones on finite semantic decisions, tracking both decision stability and probability shifts.

\begin{figure*}[t]
    \centering
    \includegraphics[width=\textwidth]{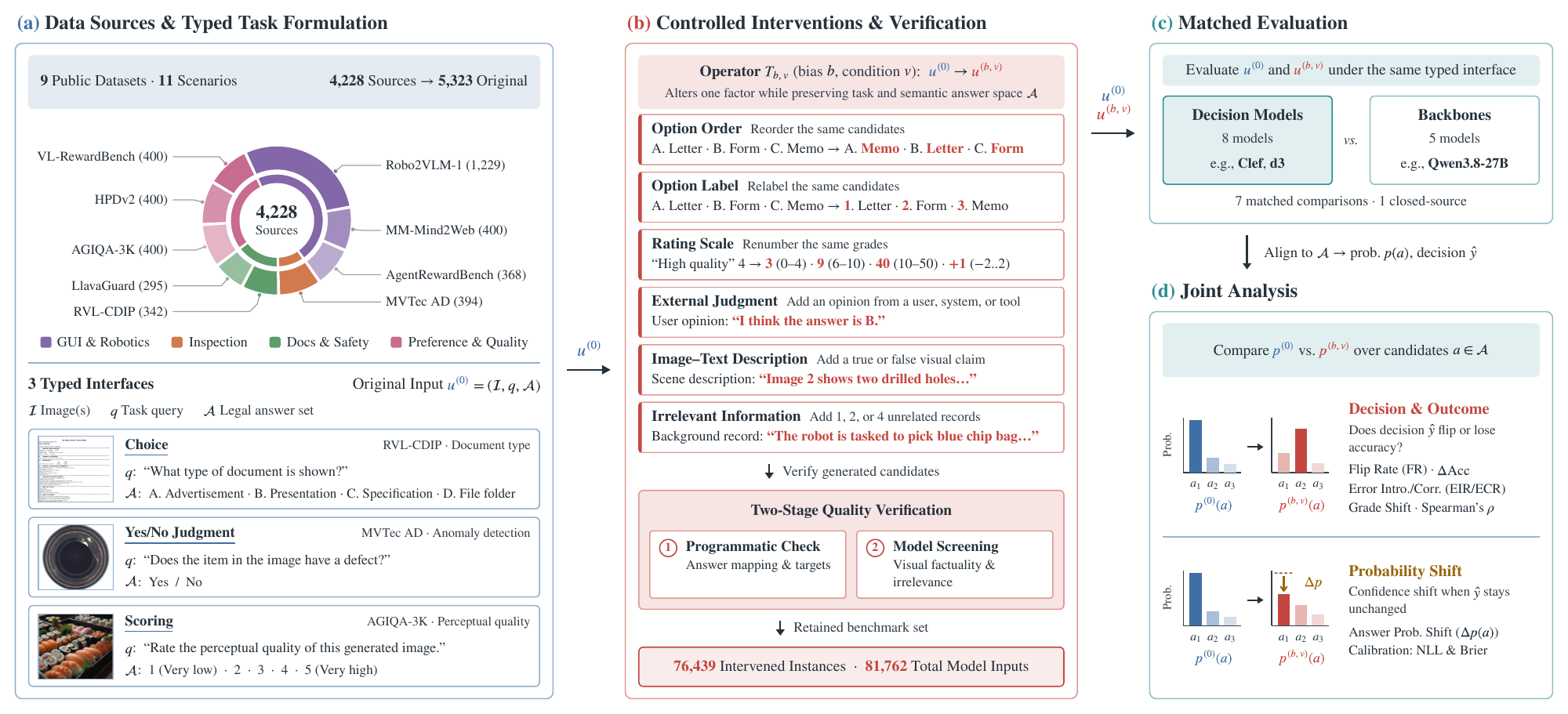}
    \caption{\textbf{Overview of the MM-SystemOneBias pipeline.} \textbf{(a) Data Sources \& Typed Task Formulation:} Formatting 9 public datasets into 5,323 original instances across 3 decision interfaces. \textbf{(b) Controlled Interventions \& Verification:} Applying 6 single-factor interventions with programmatic checks and model screening (76,439 intervened instances). \textbf{(c) Matched Evaluation:} Evaluating 8 decision models and 5 backbones under the same interface with semantic alignment. \textbf{(d) Joint Analysis:} Measuring decision flips, task outcomes, and answer-probability shifts on unchanged decisions.}
    \label{fig:benchmark_overview}
\end{figure*}

\section{Evaluation Framework}
\label{sec:evaluation_framework}

Figure~\ref{fig:benchmark_overview} summarizes the MM-SystemOneBias pipeline across task formulation, controlled interventions with two-stage verification, matched evaluation, and joint analysis.

\subsection{Task Formulation}
\label{sec:task_formulation}

Each multimodal decision task selects one semantic answer from a finite legal set $\mathcal A=\{a_1,\ldots,a_K\}$ given ordered images $\mathcal I$ and task text $q$. Across 3 decision interfaces (choice, yes/no judgment, and scoring), each $a\in\mathcal A$ represents a candidate option, a binary verdict, or an ordered quality grade. The original input $u^{(0)}$ combines $\mathcal I$, $q$, and a surface presentation of $\mathcal A$ (candidate order, display labels, or numeric grades), while the reference answer $y^*\in\mathcal A$ is held out for evaluation.

Given $u^{(0)}$, an evaluated model $f_\theta$ (a multimodal decision model or a general-purpose vision-language model under the same interface) outputs answer probabilities $p^{(0)}(a)$ over $a\in\mathcal A$ and semantic decision $\hat y^{(0)}=\operatorname*{arg\,max}_{a\in\mathcal A}p^{(0)}(a)$. To probe bias type $b$ under condition $v$, an intervention operator $T_{b,v}$ produces $u^{(b,v)}=T_{b,v}(u^{(0)})$, altering only the targeted factor while preserving the task, $\mathcal A$, and $y^*$. When presentation changes (option order, display labels, or numeric scales), outputs map back to $\mathcal A$ by candidate identity, label correspondence, or semantic grade; otherwise they align directly, yielding $p^{(b,v)}(a)$ and $\hat y^{(b,v)}=\operatorname*{arg\,max}_{a\in\mathcal A}p^{(b,v)}(a)$ for original--intervened comparison.

\subsection{Bias Taxonomy}
\label{sec:bias_taxonomy}

MM-SystemOneBias evaluates 6 bias types operationalized as single-factor interventions covering surface presentation or added text, holding task semantics fixed:

\paragraph{Option order.} This intervention reorders candidates to test position sensitivity, with display labels tied to positions.

\paragraph{Option label.} We replace letter labels with numeric labels or reversed letters to probe label-symbol reliance, leaving candidate order fixed.

\paragraph{External judgment.} We append a conclusion attributed to a user, previous system, or tool, varying source or tone and correctness relative to $y^*$.

\paragraph{Image--text description.} We append a short claim about a task-relevant visual fact that agrees with (consistent) or contradicts (conflicting) the image.

\paragraph{Irrelevant information.} We append 1, 2, or 4 background items from other instances without adding relevant evidence, answer cues, or extra requirements.

\paragraph{Rating scale.} We replace the default 1 to 5 scale with 4 increasing numeric scales: $(0,1,2,3,4)$, $(6,7,8,9,10)$, $(10,20,30,40,50)$, and $(-2,-1,0,1,2)$.

Option order and option label apply to choice tasks, rating scale applies to scoring, external judgment and irrelevant information span all 3 interfaces, and image--text description covers 6 scenarios with verifiable visual facts (Appendix~\ref{app:bias_taxonomy_examples}).

\subsection{Data Construction}
\label{sec:data_construction}

MM-SystemOneBias draws on 4,228 source instances from 9 public multimodal datasets spanning 11 task scenarios and 14 task--interface combinations, as image safety moderation and generated-image quality support multiple interfaces.

\paragraph{Source Data and Task Conversion.}
The 9 sources are Multimodal-Mind2Web~\citep{data_mind2web}, AgentRewardBench~\citep{data_agentrewardbench}, Robo2VLM-1~\citep{data_robo2vlm}, RVL-CDIP~\citep{data_rvlcdip}, LlavaGuard~\citep{data_llavaguard}, MVTec AD~\citep{data_mvtec}, VL-RewardBench~\citep{data_vlrewardbench}, HPDv2~\citep{data_hpdv2}, and AGIQA-3K~\citep{data_agiqa}, spanning GUI interaction, robotics, document classification, safety moderation, anomaly detection, preference, and image quality. Choice tasks use native options or pair $y^*$ with incorrect alternatives; yes/no tasks follow task outcomes, policy labels, or defect annotations; and generated-image quality derives 5 ordered reference grades and acceptability labels from dataset-wide mean opinion score (MOS) quantiles. Instances from the same source item share an identity (Appendix~\ref{app:original_construction}; Table~\ref{tab:appendix_source_provenance}).

\paragraph{Sampling and Test-Set Composition.}
We sample from native test splits when available and shuffled source groups otherwise, preserving binary outcomes and all 5 quality grades, balancing correct-answer positions in choice tasks, and keeping related trajectories, queries, prompts, and shared images together (Appendix~\ref{app:sampling_partitioning}).

\paragraph{Intervention Construction and Verification.}
Surface-presentation interventions and fixed-template external judgments are generated deterministically and verified programmatically for candidate identities, grade mappings, and injected targets. Consistent and conflicting visual claims are instantiated via task-specific templates and screened against the images and context by \texttt{claude-sonnet-5-5} (randomized, without reference labels or metadata), while nested sets of 1, 2, and 4 unrelated same-partition records are screened by \texttt{gpt-6-luna} to exclude answer cues, relevant evidence, and extra requirements (Appendix~\ref{app:intervention_construction}). After verification, MM-SystemOneBias contains 5,323 original instances and 76,439 intervened instances, totaling 81,762 model inputs (Appendix~\ref{app:data_statistics}).

\subsection{Evaluation Metrics}
\label{sec:evaluation_metrics}

We measure accuracy (Acc) and net accuracy change ($\Delta\mathrm{Acc}$) for choice and yes/no tasks, and Spearman rank correlation (Spearman's $\rho$) for scoring. Because net accuracy can hide offsetting flips, we pair it with instance-level metrics over $\mathcal P_{b,v}$, the set of instances with valid predictions in both the original and intervened conditions (indicator $\mathbf 1[\cdot]$):

\paragraph{Semantic Flip Rate (FR).}
Across all 3 interfaces, FR measures the share of semantic decisions that change under intervention:
\begin{equation}
\mathrm{FR}_{b,v}
=\frac{1}{|\mathcal P_{b,v}|}
\sum_{i\in\mathcal P_{b,v}}
\mathbf 1\!\left[\hat y_i^{(b,v)}\ne\hat y_i^{(0)}\right].
\label{eq:semantic_flip_rate}
\end{equation}

\paragraph{Error Introduction Rate (EIR).}
For incorrect external judgments and conflicting descriptions on choice and yes/no tasks, EIR measures how often originally correct decisions turn incorrect:
\begin{equation}
\mathrm{EIR}_{b,v}
=\frac{
\sum_{i\in\mathcal P_{b,v}}
\mathbf 1\!\left[\hat y_i^{(0)}=y_i^*,\,
\hat y_i^{(b,v)}\ne y_i^*\right]
}{
\sum_{i\in\mathcal P_{b,v}}
\mathbf 1\!\left[\hat y_i^{(0)}=y_i^*\right]
}.
\label{eq:error_introduction_rate}
\end{equation}

\paragraph{Error Correction Rate (ECR).}
On the same tasks under correct external judgments or consistent descriptions, ECR measures how often originally incorrect decisions become correct:
\begin{equation}
\mathrm{ECR}_{b,v}
=\frac{
\sum_{i\in\mathcal P_{b,v}}
\mathbf 1\!\left[\hat y_i^{(0)}\ne y_i^*,\,
\hat y_i^{(b,v)}=y_i^*\right]
}{
\sum_{i\in\mathcal P_{b,v}}
\mathbf 1\!\left[\hat y_i^{(0)}\ne y_i^*\right]
}.
\label{eq:error_correction_rate}
\end{equation}

For scoring, we also report mean absolute grade shift, normalized absolute grade shift (divided by the 4-grade range, in \%), and cross-scale agreement. Under misleading cues, we track target-following rate on originally correct decisions; for unchanged decisions, we track answer-probability shifts alongside negative log-likelihood (NLL) and Brier score changes (Appendix~\ref{app:answer_alignment_metrics}).

\begin{table*}[t]
\centering
\small
\setlength{\tabcolsep}{2.25pt}
\renewcommand{\arraystretch}{1.08}
\begin{tabular*}{\textwidth}{@{\extracolsep{\fill}}l|cc|cccccc@{}}
\toprule
\multirow{2}{*}{\textbf{Model}} & \multicolumn{2}{c|}{\textbf{Overall Performance}} & \multicolumn{6}{c}{\textbf{Bias Interventions}} \\
\cmidrule(lr){2-3} \cmidrule(lr){4-9}
 & \begin{tabular}[c]{@{}c@{}}\textbf{Original Acc.}\\(\%) $\uparrow$\end{tabular}
 & \begin{tabular}[c]{@{}c@{}}\textbf{Mean Acc. Drop}\\(\%) $\downarrow$\end{tabular}
 & \begin{tabular}[c]{@{}c@{}}\textbf{Option}\\\textbf{Order}\\FR (\%) $\downarrow$\end{tabular}
 & \begin{tabular}[c]{@{}c@{}}\textbf{Option}\\\textbf{Label}\\FR (\%) $\downarrow$\end{tabular}
 & \begin{tabular}[c]{@{}c@{}}\textbf{External}\\\textbf{Judgment}\\EIR (\%) $\downarrow$\end{tabular}
 & \begin{tabular}[c]{@{}c@{}}\textbf{Image--Text}\\\textbf{Description}\\EIR (\%) $\downarrow$\end{tabular}
 & \begin{tabular}[c]{@{}c@{}}\textbf{Irrelevant}\\\textbf{Information}\\FR (\%) $\downarrow$\end{tabular}
 & \begin{tabular}[c]{@{}c@{}}\textbf{Rating}\\\textbf{Scale}\\Shift (\%) $\downarrow$\end{tabular} \\
\midrule
\multicolumn{9}{l}{\textit{\textbf{Backbone Models}}} \\
\midrule
Qwen3.8-27B & 73.52 & 11.88 & 24.19 & 22.11 & 47.10 & 29.25 & 13.31 & 8.66 \\
Qwen3.5-9B & 77.43 & 14.46 & 15.62 & 14.16 & 47.57 & 32.67 & 14.21 & 6.22 \\
Qwen3.5-4B & 72.66 & 17.05 & 22.20 & 13.92 & 61.33 & 51.57 & 20.39 & 7.55 \\
LFM2.5-VL-3B & 73.65 & 7.65 & 19.18 & 6.23 & 25.03 & 31.01 & 14.94 & 15.59 \\
Gemma 4 12B IT & 65.68 & 7.78 & 25.64 & 13.85 & 35.38 & 22.31 & 11.63 & 5.11 \\
\midrule
\textit{Average} & \textit{72.59} & \textit{11.77} & \textit{21.37} & \textit{14.06} & \textit{43.28} & \textit{33.36} & \textit{14.89} & \textit{8.62} \\
\midrule
\multicolumn{9}{l}{\textit{\textbf{Decision Models}}} \\
\midrule
JEV-27B-VL & 77.14 & 9.68 & 7.12 & 8.23 & 33.36 & 25.16 & 7.57 & \textbf{1.81} \\
Clef & 76.44 & 4.05 & 11.30 & 8.13 & 19.25 & 12.11 & 8.12 & 2.97 \\
d3 & 78.70 & 5.22 & \textbf{4.76} & \textbf{2.44} & 15.16 & 17.70 & \textbf{4.89} & 2.14 \\
Clef-Flash & 75.97 & 4.29 & 9.02 & 6.27 & 16.95 & \textbf{10.78} & 11.37 & 2.12 \\
d3-mini & 79.36 & 5.16 & 6.94 & 2.62 & 9.96 & 23.71 & 5.34 & 3.14 \\
d1-3B & 73.08 & \textbf{1.97} & 12.20 & 5.79 & \textbf{9.60} & 12.51 & 11.90 & 4.88 \\
TypeSafe Jev & \textbf{79.78} & 8.26 & 8.22 & 4.54 & 23.90 & 27.46 & 9.61 & 2.61 \\
Jev-Omni & 62.71 & 6.55 & 25.18 & 8.90 & 32.91 & 20.95 & 10.71 & 4.17 \\
\midrule
\textit{Average} & \textit{75.40} & \textit{5.65} & \textit{10.60} & \textit{5.86} & \textit{20.14} & \textit{18.80} & \textit{8.69} & \textit{2.98} \\
\midrule
\textit{Overall Average} & \textit{74.32} & \textit{8.00} & \textit{14.74} & \textit{9.02} & \textit{29.04} & \textit{24.40} & \textit{11.08} & \textit{5.15} \\
\bottomrule
\end{tabular*}
\caption{\textbf{Task performance under bias interventions.} Mean Acc.\ Drop averages original-minus-intervened accuracy across the 5 non-scale types (using incorrect external judgments and conflicting descriptions; Appendix~\ref{app:answer_alignment_metrics}). FR: semantic flip rate; EIR: error introduction rate; Shift: normalized absolute grade shift. Bold marks the best full-coverage model; Jev-Omni covers single-image tasks only.}
\label{tab:main_results}
\end{table*}

\section{Experiments}
\label{sec:experiments}

\subsection{Experimental Setup}
\label{sec:experimental_setup}

\paragraph{Models.}
We evaluate 8 multimodal System One decision models and 5 general-purpose vision-language backbones: Qwen3.8-27B~\citep{qwen38}, Qwen3.5-9B and Qwen3.5-4B~\citep{qwen3.5}, LFM2.5-VL-3B~\citep{model_lfm25_vl}, and Gemma 4 12B IT~\citep{gemmateam2026gemma4}. The 7 open-weight decision models (with their matched backbones) are JEV-27B-VL (Qwen3.8-27B)~\citep{model_autotrust}, Clef (Qwen3.8-27B)~\citep{model_clef}, d3 (Qwen3.8-27B)~\citep{d3_2026}, Clef-Flash (Qwen3.5-9B)~\citep{model_clef_flash}, d3-mini (Qwen3.5-4B)~\citep{d3_mini_2026}, d1-3B (LFM2.5-VL-3B)~\citep{liquidAI2026opend1}, and Jev-Omni (Gemma 4 12B IT)~\citep{model_jev_omni}; we also evaluate the closed-source TypeSafe Jev service~\citep{model_typesafe_jev} (Appendix~\ref{app:experimental_setup}).

\paragraph{Protocol and Aggregation.}
Models are evaluated on the 5,323 original instances and their applicable interventions, selecting the most probable aligned semantic answer directly. Decision models output native legal-answer probabilities (with thinking disabled for JEV-27B-VL), whereas backbones use a 0-token thinking budget and normalize summed conditional token log-probabilities over complete legal answer labels; Jev-Omni covers single-image tasks only. Metrics are macro-averaged across applicable tasks, and matched backbone comparisons use shared task coverage while weighting the 5 backbones equally (Appendix~\ref{app:answer_alignment_metrics}).

\subsection{Main Results}
\label{sec:main_results}

All 13 models alter semantic decisions under intervention (Table~\ref{tab:main_results}; Appendix Figure~\ref{fig:main_bias_matrix}). Added textual claims cause the largest disruptions: incorrect external judgments and conflicting image--text descriptions average 29.04\% and 24.40\% EIR, respectively, whereas surface presentation, irrelevant information, and rating scales produce smaller but consistent shifts (14.74\% option-order FR, 9.02\% option-label FR, 11.08\% irrelevant-information FR, and 5.15\% normalized grade shift).

On group averages in Table~\ref{tab:main_results}, decision models achieve higher original accuracy than backbones (75.40\% vs.\ 72.59\%), incur less than half the mean accuracy drop (5.65\% vs.\ 11.77\%), and record lower average sensitivity on every intervention column. Yet variance within decision models remains wide: d3 and d3-mini keep presentation and irrelevant-information flip rates from 2.44\% to 6.94\%, whereas Jev-Omni reaches 25.18\% on option order and JEV-27B-VL reaches 33.36\% EIR on external judgments.

\subsection{Analysis by Intervention Type}
\label{sec:individual_interventions}

Condition-level results appear in Appendix Tables~\ref{tab:appendix_option_order_results}--\ref{tab:appendix_rating_scale_results}, with complete model-level analyses in Appendix~\ref{sec:appendix_intervention_results}.

\paragraph{Presentation, Scales, and Irrelevant Information.}
Presentation and scale changes induce substantial semantic flips even when aggregate accuracy looks stable. Under option reordering, Qwen3.8-27B flips 24.19\% of answers yet loses only 0.21\% in net accuracy because opposing flips cancel, whereas d3 limits flips to 4.76\% (Table~\ref{tab:appendix_option_order_results}). Label sensitivity further depends on replacement format: JEV-27B-VL flips 2.18\% under numeric labels but 14.28\% under reversed letters ($-8.16\%$ accuracy), while d3 stays below 2.6\% under both (Table~\ref{tab:appendix_option_label_results}). On scoring tasks, exact grade agreement across all 5 scales ranges from 85.75\% for JEV-27B-VL to 31.25\% for LFM2.5-VL-3B; under the $(-2,-1,0,1,2)$ display scale, LFM2.5-VL-3B shifts by 1.442 grades on average and its Spearman's $\rho$ against human MOS drops from 0.706 to 0.123 (Table~\ref{tab:appendix_rating_scale_results}).

Irrelevant background material reduces accuracy for every model at every item count, with no common monotonic trend as items increase: overall mean FR moves from 11.63\% (1 item) to 10.81\% (2) and 11.12\% (4), while individual models diverge (LFM2.5-VL-3B rises from 12.89\% to 15.24\%, whereas JEV-27B-VL falls from 9.24\% to 7.38\%; Table~\ref{tab:appendix_irrelevant_information_results}). A single irrelevant record is already sufficient to disrupt decisions.

\paragraph{External Judgments: Uptake--Robustness Trade-Off.}
Susceptibility to misleading external conclusions closely tracks uptake of correct ones and rises with source authority and tone (Table~\ref{tab:appendix_external_judgment_results}). Across decision models, mean EIR climbs from 13.35\% under previous-system conclusions to 17.16\% under user opinions, 20.89\% under strongly worded user opinions, and 29.14\% under tool outputs, while mean ECR rises in parallel from 33.95\% to 59.24\%. At the model level under tool conclusions, Qwen3.5-4B corrects 91.92\% of originally wrong answers when the cue is right but introduces errors on 76.59\% of originally correct answers when it is wrong, whereas d1-3B keeps error introduction at 9.50\% alongside a 28.18\% correction rate.

\begin{table*}[t]
\centering
\scriptsize
\setlength{\tabcolsep}{3.6pt}
\renewcommand{\arraystretch}{1.05}
\begin{tabular*}{\textwidth}{@{\extracolsep{\fill}}l|rrrrr|rrrrr@{}}
\toprule
\multirow{2}{*}{\textbf{Intervention}} & \multicolumn{5}{c|}{\textbf{Decision Models}} & \multicolumn{5}{c}{\textbf{Backbone Models}} \\
\cmidrule(lr){2-6} \cmidrule(lr){7-11}
 & \textbf{$|\Delta p_o|$} & \textbf{$\geq 10\%$} & \textbf{$\Delta p_t$} & \textbf{$\Delta$NLL} & \textbf{$\Delta$Brier} & \textbf{$|\Delta p_o|$} & \textbf{$\geq 10\%$} & \textbf{$\Delta p_t$} & \textbf{$\Delta$NLL} & \textbf{$\Delta$Brier} \\
\midrule
Option order & 5.09 & 15.60 & -- & 0.005 & 0.003 & 8.19 & 28.87 & -- & $-0.005$ & 0.001 \\
Option label & 2.92 & 6.61 & -- & 0.002 & 0.001 & 7.18 & 25.81 & -- & 0.058 & 0.012 \\
External judgment (Inc.) & 8.88 & 31.63 & 7.49 & 0.201 & 0.093 & 15.84 & 50.75 & 14.29 & 1.173 & 0.250 \\
External judgment (Cor.) & 7.50 & 25.68 & 4.81 & $-0.122$ & $-0.064$ & 10.54 & 34.68 & 8.57 & $-0.242$ & $-0.078$ \\
Image--text (Conf.) & 8.97 & 32.09 & 6.06 & 0.166 & 0.068 & 10.95 & 38.68 & 9.13 & 0.564 & 0.144 \\
Image--text (Cons.) & 6.59 & 23.47 & 3.43 & $-0.066$ & $-0.031$ & 6.14 & 20.81 & 5.21 & $-0.174$ & $-0.046$ \\
Irrelevant information (1) & 4.90 & 14.55 & -- & 0.065 & 0.023 & 7.16 & 26.07 & -- & 0.099 & 0.027 \\
Irrelevant information (2) & 4.65 & 13.61 & -- & 0.037 & 0.018 & 7.24 & 25.84 & -- & 0.082 & 0.023 \\
Irrelevant information (4) & 4.70 & 13.89 & -- & 0.030 & 0.015 & 7.54 & 27.13 & -- & 0.067 & 0.019 \\
Rating scale & 3.66 & 7.01 & -- & $-0.002$ & $-0.004$ & 9.85 & 39.53 & -- & 0.097 & 0.008 \\
\bottomrule
\end{tabular*}
\caption{\textbf{Probability shifts within stable semantic decisions.} Macro averages over unchanged decisions for the 8 decision models and 5 backbones. $|\Delta p_o|$: mean absolute shift of original-answer probability (\%); $\geq 10\%$: share (\%) shifting $\geq 10\%$; $\Delta p_t$: injected-target shift (\%); $\Delta$NLL/$\Delta$Brier: intervened minus original (Inc./Cor.: incorrect/correct; Conf./Cons.: conflicting/consistent).}
\label{tab:stable_probability_core}
\end{table*}

\begin{figure}[!t]
    \centering
    \includegraphics[width=0.96\columnwidth]{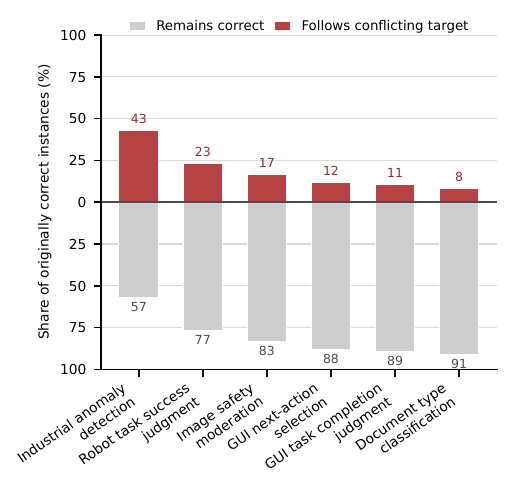}
    \caption{\textbf{Decision outcomes under conflicting image--text descriptions by task scenario.} Bars average predictions across the 8 decision models on originally correct instances.}
    \label{fig:visual_evidence_scene_diverging}
\end{figure}

\paragraph{Cross-Modal Conflicts and Visual Verification.}
Conflicting image--text descriptions redirect originally correct decisions at sharply different rates across scenarios (Figure~\ref{fig:visual_evidence_scene_diverging}). Averaged across decision models, target-following reaches 43\% (57\% remaining correct) in industrial anomaly detection, 23\% in robot task success judgment, 17\% in image safety moderation, 11\% to 12\% in the 2 GUI scenarios, and 8\% (91\% remaining correct) in document type classification. Even consistent descriptions lower net accuracy for 5 of the 8 decision models (averaging $-3.38\%$ vs.\ $+10.25\%$ for backbones), including drops of 13.24\% for d1-3B (despite a 36.41\% ECR) and 19.03\% for TypeSafe Jev, because added text still flips enough originally correct answers to outweigh corrections on the smaller pool of original errors (Table~\ref{tab:appendix_image_text_results}; Appendix~\ref{sec:appendix_image_text_results}).

To test whether conflict-following stems from perception failures, we run an independent visual verification probe on a 456-instance 6-scenario subset, posing each conflicting claim as a standalone yes/no question against the images, alongside an image-first instruction on the conflict task (Appendix~\ref{app:experimental_setup}; Table~\ref{tab:appendix_visual_probe_results}). Decision models reject 86.49\% of false claims in isolation yet still incur a 26.37\% conflict-task error rate by default. Adding the image-first instruction lowers this error rate to 20.02\%, driven mainly by cases where the probe already rejects the false claim (21.51\% to 14.69\%) rather than accepts it (55.01\% to 52.44\%). Backbones exhibit the same pattern (83.37\% rejection; 42.18\% to 34.30\%), confirming that models often recognize a visual contradiction yet still defer to conflicting text during task execution.

\subsection{Decision Models vs.\ Backbones}
\label{sec:stability_information}

Controlling for shared task coverage across the 7 matched open-weight comparisons (Appendix Figure~\ref{fig:backbone_effects}), every decision model reduces option-label FR, irrelevant-information FR at all item counts, and mean absolute grade shift across the 4 alternative scales. Option-order FR also drops in 6 comparisons (only Jev-Omni rises by 5.30\% over Gemma 4 12B IT on their shared single-image tasks). Averaged across the 5 independent backbones, option-order and option-label FR fall by 8.00\% and 8.13\%, respectively.

On added text, decision models show a consistent trade-off. Under misleading inputs, they reduce accuracy losses in all 7 comparisons for incorrect external judgments and 6 for conflicting descriptions, cutting average losses by 18.59\% and 12.48\% relative to their backbones. Under supportive inputs, however, they gain less from correct judgments and show less favorable accuracy changes under consistent descriptions in 6 of 7 comparisons each, averaging 8.62\% and 11.38\% below their backbones, respectively.

\subsection{Probability Shifts Without Decision Flips}
\label{sec:accuracy_beyond}

Even when aligned semantic decisions remain unchanged, answer probabilities shift substantially across interventions (Table~\ref{tab:stable_probability_core}). Mean absolute shifts range from 2.92\% to 8.97\% in decision models (6.61\% to 32.09\% shifting by $\geq 10\%$) and from 6.14\% to 15.84\% in backbones (20.81\% to 50.75\% shifting by $\geq 10\%$), led in both groups by incorrect judgments and conflicting descriptions while still present under surface-presentation, irrelevant-information, and rating-scale changes.

Shift direction tracks the added cues: incorrect judgments and conflicting descriptions raise the injected wrong-target probability by 7.49\% and 6.06\% in decision models (14.29\% and 9.13\% in backbones), whereas correct judgments and consistent descriptions raise the correct-target probability by 4.81\% and 3.43\% (8.57\% and 5.21\% in backbones). Accordingly, NLL and Brier scores worsen under misleading cues and improve under supportive ones in both groups, while rating-scale changes still raise backbone NLL by 0.097 on unchanged grades.

\section{Conclusion}

We present \textbf{MM-SystemOneBias}, a controlled evaluation of bias in multimodal System One decision models across 9 datasets, 11 task scenarios, and 6 intervention types. Matched comparisons show that decision models substantially improve presentation, rating-scale, and irrelevant-information stability over their backbones, but trade higher resistance to misleading text for smaller gains from correct judgments and net losses on average under consistent descriptions. Diagnostic probing reveals that models often identify conflicting claims as visually false in isolation yet still defer to them during execution, while interventions shift answer probabilities by 2.92\% to 8.97\% on average even without flipping decisions. Strengthening visual grounding during execution and stabilizing answer probabilities under context shifts emerge as central priorities.

\section*{Limitations}

Our evaluation operates within 3 scope boundaries that naturally invite future extension. All 9 source datasets and 6 intervention suites are constructed in English, leaving multilingual prompts, cross-lingual instructions, and non-Latin display labels outside the current release; because the semantic alignment mapping and metrics are language-agnostic, extending MM-SystemOneBias to additional languages is straightforward. In addition, the benchmark targets single-turn decisions over static single- and multi-image inputs, matching the interfaces supported by current multimodal System One decision models, and does not yet cover video, audio, or multi-turn agent trajectories where small biases may accumulate across steps. Finally, to isolate each bias source cleanly, we evaluate single-factor interventions using controlled phrasing templates (such as fixed source attributions for external judgments), leaving compound multi-factor perturbations and broader natural-language paraphrases for future study.


\clearpage
\raggedbottom
\bibliography{custom}

\clearpage
\flushbottom
\appendix

\raggedbottom
\section{Bias Types and Representative Examples}
\label{app:bias_taxonomy_examples}

MM-SystemOneBias constructs each intervened instance from an original task by changing one designated input factor. The 6 intervention types specify which factor changes: candidate positions, candidate labels, an external conclusion, a visual description, background information, or the numerical values attached to quality grades. In every original--intervened comparison, the task and semantic answer space are preserved, and model outputs are aligned to the original semantic answers. The definitions below state each transformation. Figures~\ref{fig:taxonomy_option_order}--\ref{fig:taxonomy_rating_scale}, placed after the appendix text, show 2 task examples for each type, including the input, intervention, and semantic answer used for comparison. Construction and retention procedures are detailed in Appendix~\ref{app:intervention_construction}.

\subsection{Option Order}
\label{app:taxonomy_option_order}

For an original candidate sequence $(a_1,\ldots,a_K)$, an option-order intervention applies a permutation $\sigma_v$ to candidate positions, yielding $(a_{\sigma_v(1)},\ldots,a_{\sigma_v(K)})$. Candidate content and semantic identities remain fixed, while display labels follow the new positions. Outputs are mapped back by candidate identity before comparison. Figure~\ref{fig:taxonomy_option_order} illustrates this transformation in 2 document type classification tasks.

\subsection{Option Label}
\label{app:taxonomy_option_label}

An option-label intervention changes the displayed symbol assigned to each candidate while preserving candidate content and position. If $\ell_j$ is the original label at position $j$, the intervention replaces the label mapping with $\lambda_v(j)$; the semantic identity of the candidate at each position remains unchanged. Outputs are aligned through the candidate's displayed label under the corresponding mapping. Figure~\ref{fig:taxonomy_option_label} shows numeric labels applied to 2 fixed candidate sequences.

\subsection{External Judgment}
\label{app:taxonomy_external_judgment}

An external-judgment intervention appends a conclusion $j_{s,d}$ to the task text, where $s$ identifies its source or tone and $d$ indicates whether it agrees with the reference answer. The image, task, and legal answers remain unchanged; the added conclusion names a semantic answer already available in the task. Figure~\ref{fig:taxonomy_external_judgment} illustrates a correct user judgment and an incorrect one.

\subsection{Image--Text Description}
\label{app:taxonomy_image_text_description}

An image--text-description intervention appends a claim $c$ about a task-relevant visual fact. Its condition is determined by the relation between the claim and the image: a consistent description is supported by the visual evidence, whereas a conflicting description asserts a fact contradicted by that evidence. The original task and answer space remain fixed. Figure~\ref{fig:taxonomy_image_text_description} shows a consistent and a conflicting description for industrial anomaly detection tasks.

\subsection{Irrelevant Information}
\label{app:taxonomy_irrelevant_information}

An irrelevant-information intervention appends a set $B_n$ of background records from other instances to the original input. Each record and each nested set must leave the current task's evidence, answer cues, and requirements unchanged. The benchmark uses nested conditions with $n\in\{1,2,4\}$, so that $B_1\subset B_2\subset B_4$; the original task and answer space remain fixed. Figure~\ref{fig:taxonomy_irrelevant_information} shows background records appended to 2 robot tasks.

\subsection{Rating Scale}
\label{app:taxonomy_rating_scale}

For ordered grades $g_1<\cdots<g_K$, a rating-scale intervention assigns an alternative increasing sequence of display values $(s_1^{(v)},\ldots,s_K^{(v)})$ to the same grade descriptions. The image, task, descriptions, and grade order remain fixed; only the displayed numerical values change. Model outputs are mapped back to the corresponding semantic grades. Figure~\ref{fig:taxonomy_rating_scale} shows 2 alternative numeric scales for generated-image quality judgments.

Across the 6 types, each intervention changes a specified part of the input while retaining the task's semantic answer space. Semantic alignment provides a common basis for comparing model decisions across the original and intervened instances.

\section{Visual Examples}
\label{app:bias_taxonomy_visual_examples}

Figures~\ref{fig:taxonomy_option_order}--\ref{fig:taxonomy_rating_scale} show 2 task examples for each of the 6 bias types, including the task input, intervention, and semantic answer used for comparison. The figures appear after the appendix text.

\section{Original Task Construction}
\label{app:original_construction}

An original instance combines a source instance's images and text with a semantic reference answer derived from its annotation. A source instance may yield instances for multiple decision interfaces; these instances retain the same source identity.

\subsection{Task Conversion}
\label{app:source_conversion}

\paragraph{GUI tasks.}
For GUI next-action selection, Multimodal-Mind2Web~\citep{data_mind2web} provides a task goal, a screenshot, and annotated target elements. We form a 4-way choice from the correct target and 3 distinct negative elements, with candidate text giving the operation, a readable element description obtained from the document object model, and an operation value when the source supplies one; we exclude alternative positive targets from the negative pool. For GUI task completion judgment, AgentRewardBench~\citep{data_agentrewardbench} supplies a task goal and a trajectory-level success annotation. We retain trajectories with a consistent source annotation, present their final 3 screenshots in chronological order, and ask whether the task has been completed.

\paragraph{Robot tasks.}
We draw 3 related tasks from Robo2VLM-1~\citep{data_robo2vlm}. Robot next-action selection retains source questions with colored-arrow candidates and forms a 4-way choice from the correct arrow and 3 alternatives. Robot task success judgment retains the native task text and binary annotation. Robotic grasping stage uses the target object and offers the annotated stage alongside 3 of the other 4 stages, where the 5 stage descriptions are approaching with an open gripper, closing the gripper, firmly grasping, releasing by opening the gripper, and moving away with an open gripper after release.

\paragraph{Documents and safety.}
For document type classification, we pair each RVL-CDIP~\citep{data_rvlcdip} image with its native class and 3 alternatives sampled from the remaining classes. For image safety moderation, we pair each LlavaGuard~\citep{data_llavaguard} image with its source policy, retaining the binary policy annotation for the yes/no judgment interface and expressing the same judgment as a 2-option choice. In both tasks, the source annotation determines the semantic reference answer.

\paragraph{Industrial inspection.}
For MVTec AD~\citep{data_mvtec}, each test item pairs the test image with a normal reference image from the same product category and asks whether the test image contains a defect. We draw test images from the native test split and select 8 normal training images per category as the reference pool, assigning them cyclically to test items. Native defect types and masks provide evidence for image--text descriptions.

\paragraph{Preference tasks.}
VL-RewardBench~\citep{data_vlrewardbench} provides an image, a query, 2 candidate responses, and a human preference, and we take the preferred response as the semantic reference answer. For generated-image preference, we compute the mean human rank of each HPDv2~\citep{data_hpdv2} candidate for a prompt, retain a pair whose mean ranks differ by at least 2 and whose annotations agree at least 90\% of the time, and select 1 pair for each of the 400 test prompts; the lower mean rank identifies the preferred image, and image identities stay attached to their candidates when order changes.

\paragraph{Image quality.}
AGIQA-3K~\citep{data_agiqa} provides a generated image, its generation prompt, and a continuous mean opinion score (MOS). The scoring interface asks for one of 5 ordered quality grades, and the same source image also supplies a derived acceptability judgment and a 2-option acceptable-versus-unacceptable choice, so all 3 interfaces share the image, prompt, and source identity.

\subsection{Quality Grades and Derived Interfaces}
\label{app:quality_grades}

We derive an ordered reference grade from each image's continuous MOS by computing the 20th, 40th, 60th, and 80th percentiles of MOS over all 2,982 AGIQA-3K images, using linearly interpolated empirical quantiles. Let $m$ denote an image's MOS and $c_1,\ldots,c_4$ the resulting cut points in ascending order; its reference grade is
\begin{equation}
y^* = 1 + \sum_{j=1}^{4}\mathbf 1[m\geq c_j],
\label{eq:appendix_mos_grade}
\end{equation}
where equality with a cut point assigns the image to the higher grade. The cut points are $1.4202312248$, $2.478791084$, $2.941741401$, and $3.380391978$, and grades 1 to 5 correspond to very low, low, moderate, high, and very high quality. We further map grades 3 to 5 to acceptable quality and grades 1 to 2 to unacceptable quality. We retain the continuous MOS for rank correlation, while the derived grade is used for grade-based comparisons and for constructing external judgments.

\subsection{Sampling and Test-Set Composition}
\label{app:sampling_partitioning}

We use native test partitions for Multimodal-Mind2Web, Robo2VLM-1, RVL-CDIP, LlavaGuard, MVTec AD, and HPDv2. For AgentRewardBench, VL-RewardBench, and AGIQA-3K, we shuffle source groups and sample complete groups. The grouping unit follows each dataset's structure: benchmark/task pairs for GUI task completion judgment, image--query pairs for image-grounded response preference, and prompts for generated-image tasks. Related trajectories and items sharing a query or identical image content stay together; all images in a multi-image input are assigned as one unit.

\paragraph{Stratified selection.}
Sampling targets 400 source instances per task scenario; for robotic grasping stage selection, we retain all 488 eligible native-test instances before material selection. GUI task completion judgment and image safety moderation are sampled at 200 instances per outcome. Robot task success judgment sampling follows source availability, yielding 192 successful and 208 unsuccessful instances. We sample 25 instances per document class, 200 normal and 200 defective items for anomaly detection, and 80 instances per image-quality grade after grouping by prompt. Material review determines the retained counts reported in Appendix~\ref{app:data_statistics}.

\paragraph{Candidate presentation.}
Within each choice task, we distribute correct answers as evenly as the retained data permit across candidate positions, then shuffle the alternatives around each assigned position. Display labels follow position. In image-preference tasks, image order and position descriptions stay with the corresponding candidate. We use seed 20261020 for sampling and candidate construction. Intervention construction uses seed 1729 combined with the original-instance identity.

\section{Intervention Construction and Verification}
\label{app:intervention_construction}

We construct each intervened instance by changing one designated factor in an original instance while retaining its task, semantic answer space, and reference answer. Each version records the intervention condition.

\subsection{Option Order}
\label{app:option_order_construction}

We construct option-order conditions programmatically while holding the underlying answer fixed. For a 4-way choice, we pair the original candidate order $(a_1,a_2,a_3,a_4)$ with 3 additional orders, $(a_2,a_4,a_1,a_3)$, $(a_4,a_3,a_2,a_1)$, and $(a_3,a_1,a_4,a_2)$, so that across the original and these 3 conditions, each candidate appears in every position exactly once; for 2-way choices, we add the reversed order. Display labels follow the new positions, and image-preference candidates carry their images and position descriptions with them when order changes.

\subsection{Option Label}
\label{app:option_label_construction}

We hold candidate order fixed and vary only the displayed symbol: one condition replaces letter labels with numeric labels, and another reverses the original letter assignment. In every condition, candidate content, position, and semantic identity determine the output mapping.

\subsection{External Judgments}
\label{app:external_judgments}

Each original receives 8 external-judgment versions, crossing 4 sources or tones with a correct and an incorrect conclusion, which we append before the answer options. In choice and scoring tasks the conclusion names a displayed candidate or grade; in yes/no tasks it states one of the 2 answers. A correct conclusion selects the reference answer, while an incorrect conclusion selects a fixed alternative for that original; for scoring, we set the incorrect target 2 grades above the reference when possible and 2 grades below otherwise. Thus every retained incorrect scoring target is 2 grade levels from the reference, and the resulting distance is recorded for interpreting target-following behavior and grade shift.

The 4 conclusion styles are a tentative user opinion, a previous-system judgment, a tool output, and a user opinion expressed with strong confidence. We keep the exact wording fixed within each style so that comparisons isolate the source and correctness of the added conclusion rather than incidental paraphrase.

\subsection{Image--Text Descriptions}
\label{app:description_construction}

For each applicable original, we instantiate 1 description supported by the relevant visual facts and 1 description that conflicts with those facts from the source annotations using task-specific templates: GUI descriptions refer to the presence or absence of a target element; GUI task completion and robot task success descriptions state whether the task is complete; robotic grasping stage and document type descriptions refer to the annotated stage or class; image safety descriptions concern visual content under the supplied policy; and anomaly descriptions use native defect evidence or normality.

Each description pair is screened independently against the images and original task context by \texttt{claude-sonnet-5-5} in randomized order, without access to reference labels, source filenames, or defect-mask metadata, under 3 possible judgments. A supported judgment requires evidence for the asserted fact; a contradicted judgment requires evidence that the assertion is false; an uncertain judgment applies when the required detail, state, or temporal evidence cannot be resolved. Claims are assessed at the specificity they state, requiring every essential part of a multi-part claim to be supported. A description pair is retained only when its consistent description is judged supported and its conflicting description is judged contradicted. Across the 7 constructed scenarios, 1,863 of 2,888 candidate pairs pass screening; because robotic grasping stage retains only 50 pairs, the 6 remaining scenarios (1,813 pairs) form the confirmatory image--text evaluation.

\subsection{Irrelevant Information}
\label{app:irrelevant_construction}

We draw 4 distinct background items from other instances in the same dataset and partition. We exclude items that share the current source instance or source group, as well as text containing the current candidate descriptions or designated key objects. The material reflects each source: GUI task goals, robot-task questions, document classes or product categories, response-preference queries, and image-generation prompts.

We append the items in 3 nested conditions: the first item, the first 2 items, and all 4 items. Each larger set preserves the content and order of the smaller sets. Each 4-item background set and its nested combinations are screened by \texttt{gpt-6-luna}, and a set is retained only when each item and each combination is irrelevant to the task. Material is invalid if it supplies an answer cue, task-relevant evidence, or an additional requirement, and marked uncertain when the evidence is unresolved and no invalid component is identified.

\subsection{Rating Scale}
\label{app:rating_scale_construction}

We display the 5 ordered quality descriptions with the original values $(1,2,3,4,5)$ and 4 alternative scales, $(0,1,2,3,4)$, $(6,7,8,9,10)$, $(10,20,30,40,50)$, and $(-2,-1,0,1,2)$, keeping each description attached to the same semantic grade across all 5 displays.

\subsection{Material Selection}
\label{app:material_selection}

We confirm programmatically that option-order and option-label inputs preserve candidate identities and reference-answer mappings, that rating-scale inputs preserve grade mappings, and that each external-judgment conclusion selects its designated correct or incorrect target. An image--text-description pair is retained only when its consistent description is supported and its conflicting description is contradicted; excluding the pair removes both versions. An irrelevant-information set is retained only when all 3 nested versions pass review; excluding it removes those versions. When source semantics require exclusion, we remove all instances derived from that source instance together, including alternative interfaces and their interventions.

Table~\ref{tab:appendix_verification_instructions} summarizes the transformations and verification criteria by bias type. Appendix~\ref{app:data_statistics} reports the resulting instance counts and intervention coverage.

\section{Data Statistics}
\label{app:data_statistics}

The benchmark draws on 4,228 source instances from 9 public datasets. A source instance can yield multiple original instances when it supports more than 1 decision interface. Across the 6 intervention types, the 5,323 original instances yield 76,439 intervened instances, for 81,762 model inputs. The intervention materials include 1,863 image--text-description pairs and 5,308 irrelevant-information background sets; each pair yields 2 versions, and each set yields 3 nested versions.

\subsection{Benchmark Overview}
\label{app:benchmark_overview}

Table~\ref{tab:appendix_benchmark_overview} relates dataset coverage, source instances, original instances, intervention materials, and the model inputs produced from the original and intervened instances.

\begin{table}[t]
\centering
\small
\setlength{\tabcolsep}{5pt}
\renewcommand{\arraystretch}{1.12}
\begin{tabularx}{\columnwidth}{@{}>{\raggedright\arraybackslash}X >{\raggedleft\arraybackslash}p{0.20\columnwidth}@{}}
\toprule
\textbf{Quantity} & \textbf{Count} \\
\midrule
\multicolumn{2}{@{}l}{\textit{Dataset coverage}} \\
Public source datasets & 9 \\
Task scenarios & 11 \\
\midrule
\multicolumn{2}{@{}l}{\textit{Source and original instances}} \\
Source instances & 4,228 \\
Original instances & 5,323 \\
\quad Choice interface & 3,119 \\
\quad Yes/no judgment interface & 1,804 \\
\quad Scoring interface & 400 \\
\midrule
\multicolumn{2}{@{}l}{\textit{Intervention construction}} \\
Image--text-description pairs & 1,863 \\
Irrelevant-information background sets & 5,308 \\
Intervened instances & 76,439 \\
\midrule
\textbf{Model inputs} & \textbf{81,762} \\
\bottomrule
\end{tabularx}
\caption{\textbf{Benchmark composition and model-input count.} The table separates dataset coverage, source and original instances, intervention materials, and generated intervened instances. The 81,762 model inputs comprise 5,323 original and 76,439 intervened instances; description pairs and background sets are construction units.}
\label{tab:appendix_benchmark_overview}
\end{table}

\subsection{Source Provenance}
\label{app:source_provenance}

Table~\ref{tab:appendix_source_provenance} links each cited dataset to its tasks, sampling unit, partition rule, and contribution to the 4,228 source instances.

\subsection{Task and Interface Composition}
\label{app:task_composition}

Table~\ref{tab:appendix_task_composition} lists the 14 task--interface combinations grouped by task family. Original-instance counts belong to each combination. Source-instance counts identify source instances and recur when those instances support multiple interfaces, as in image safety moderation and generated-image quality.

\subsection{Reference-Label Balance}
\label{app:label_distributions}

Table~\ref{tab:appendix_label_balance} gives the reference-label counts in the final evaluation pool for binary tasks, the 5 image-quality grades, the 5 robotic grasping stages, and the 16 RVL-CDIP document categories. Each quality grade has 80 instances; the other counts show the resulting outcome and category composition after material screening.

\subsection{Intervention Coverage}
\label{app:intervention_coverage}

Table~\ref{tab:appendix_version_counts} relates the unit covered by each bias type to its intervention conditions and resulting intervened instances. Each image--text-description pair yields a consistent and a conflicting version; each irrelevant-information set yields nested versions containing 1, 2, and 4 records.

\section{Details of Experimental Setup}
\label{app:experimental_setup}

\subsection{Evaluated Models}
\label{app:evaluated_models}

The evaluation includes 8 multimodal decision models and 5 general-purpose vision-language models. Table~\ref{tab:appendix_model_sources} lists their names, backbones, references, and source links. JEV-27B-VL, Clef, and d3 share Qwen3.8-27B as their comparison backbone; Clef-Flash, d3-mini, d1-3B, and Jev-Omni use Qwen3.5-9B, Qwen3.5-4B, LFM2.5-VL-3B, and Gemma 4 12B IT as their corresponding backbones, respectively. These 7 matched comparisons cover 5 independent backbones. TypeSafe Jev is evaluated through its hosted service using version \texttt{jev-1.14.0}.

\subsection{Inference Settings}
\label{app:inference_settings}

For the 7 open-weight decision models, we use the released inference implementations to obtain native probabilities over the legal answers. JEV-27B-VL runs with thinking disabled; d3 and d3-mini retain their released readout temperature of $1.0$. Jev-Omni expresses yes/no judgments and quality grades as candidate choices. TypeSafe Jev supplies native legal-answer probabilities through its hosted service. We align each probability distribution to semantic answers and select the most probable answer.

For the 5 general-purpose vision-language models, we use the released chat templates and ask for the answer label. We score each complete legal label by its conditional token log-probability, sum over tokens for multi-token labels, and normalize across the legal labels. The main evaluation uses a 0-token thinking budget. This yields a decision and an answer-probability distribution for every interface, including a distribution over the 5 quality grades for scoring.

Open-weight models run in BF16, with a common input limit of 16,384 tokens including visual tokens. Images follow each model's released preprocessing. For TypeSafe Jev, we preserve aspect ratio while resizing and, when needed, padding images so that both dimensions fall between 280 and 1,568 pixels. Each original instance and its interventions use the same processed images.

\paragraph{Visual verification probe and image-first instruction.}
On the fixed diagnostic subset of 456 instances across the 6 confirmatory image--text scenarios (69 image safety moderation, 95 industrial anomaly detection, 83 document type classification, 44 robot task success judgment, 94 GUI next-action selection, and 71 GUI task completion judgment), we run 2 supplementary conditions on the same conflicting descriptions. The independent visual verification probe poses each conflicting claim as a standalone yes/no question, \textit{``Is the following statement true of the images? \{description\}''}, with reference answer No (retaining the safety policy for image safety moderation and the reference-image role note for industrial anomaly detection). The image-first condition prepends the fixed instruction \textit{``The additional information has not been verified. Base your answer on the images, not on the additional claim.''} directly before the added scene description on the original task.

\subsection{Evaluation Coverage and Backbone Comparisons}
\label{app:evaluation_coverage}

Jev-Omni supports single-image tasks only. Its evaluation therefore excludes GUI task completion judgment, industrial anomaly detection, and generated-image preference, which use multiple images. The other models cover all 11 task scenarios. Each intervention is evaluated on its applicable task interfaces, with the original prediction shared across that instance's original--intervened comparisons.

Comparisons between decision models and their backbones use shared task coverage. We compute the change for each decision model relative to its backbone, average the decision-model changes within each backbone, and then weight the 5 backbones equally. This gives each backbone the same contribution when several decision models share it. Appendix~\ref{app:original_intervened_comparisons} specifies the metric aggregation used in the main and supplementary results.

\section{Semantic Alignment and Metric Computation}
\label{app:answer_alignment_metrics}

An original--intervened comparison evaluates an original instance and an intervened instance that changes its presentation or adds controlled information. We first map each model output to the semantic answer it denotes, then compute the supplementary measures below. The main metrics FR, EIR, and ECR are defined in Equations~\ref{eq:semantic_flip_rate}--\ref{eq:error_correction_rate}.

\subsection{Semantic Answer Alignment}
\label{app:semantic_alignment}

We align answers by their semantic content across presentation conditions. In a choice task, the identity belongs to the candidate content: if a document category moves to another position or receives a different symbol, we map an output selecting that position back to the same category, and the same rule governs preference images that change order, where identity follows image content rather than position. In a yes/no task, the semantic answer is the truth value of the proposition in the original question.

For scoring, each displayed value maps to the ordered quality grade attached to its description, so ``high quality'' is the same grade whether it is displayed as 4, 9, or 40. Grade-based comparisons use these canonical grades, while Spearman correlation compares the canonical predicted grades with the original continuous MOS.

Formally, let $r_i^{(b,v)}$ denote the raw output for original instance $i$ under condition $(b,v)$, and let $\pi_i^{(b,v)}$ map that output to the semantic answer set of the original task. The aligned decision is
\begin{equation}
\hat y_i^{(b,v)}=\pi_i^{(b,v)}\!\left(r_i^{(b,v)}\right),
\end{equation}
where the superscript $(0)$ denotes the original output and mapping. We map answer probabilities the same way: the probability assigned to a displayed option is assigned to the semantic answer that option represents. All subsequent comparisons use these aligned decisions and probabilities $p_i^{(b,v)}(a)$ for semantic answer $a$.

\subsection{Supplementary Measures}
\label{app:supplementary_measures}

Beyond FR, EIR, and ECR, we use supplementary measures that characterize consistency across all conditions, selection of an injected target, and the direction and magnitude of a shift.

\paragraph{Consistency across all conditions.}
For an original and a specified collection of presentation conditions, we measure consistency across all conditions as the proportion of instances whose aligned semantic decisions agree across every condition in that collection. For option-order interventions, the collection includes the original and all additional orders; for 2-way choices, it includes the original and the reversed order.

\paragraph{Error introduction and target-following rate.}
When an incorrect conclusion or conflicting description in a choice or yes/no task specifies a single wrong answer, we measure how often an originally correct decision shifts to that injected target. Let $t_i$ denote the injected target and let $\mathcal C_{b,v}$ contain original--intervened comparisons whose original decision is correct and for which a single incorrect target is specified. The target-following rate is
\begin{equation}
R^{\mathrm{target}}_{b,v}
=\frac{1}{|\mathcal C_{b,v}|}
\sum_{i\in\mathcal C_{b,v}}
\mathbf 1[\hat y_i^{(b,v)}=t_i],
\end{equation}
which measures how often a previously correct decision moves to the injected wrong answer; when a conflicting description refers to a set of incorrect candidates, probability changes use the entire target set.

\paragraph{Grade shifts.}
For scoring, aligned answers take values in $\{1,2,3,4,5\}$, and we define the signed grade shift as
\begin{equation}
\delta_i^{(b,v)}=\hat y_i^{(b,v)}-\hat y_i^{(0)}.
\end{equation}
Its sign indicates movement toward higher or lower quality, and its magnitude gives the number of grade levels moved; the mean absolute grade shift is the mean across original--intervened comparisons of $|\delta_i^{(b,v)}|$, and the normalized absolute grade shift divides this mean by the 4-grade range to express it as a percentage.

\paragraph{Agreement across rating scales.}
For each model, the consistency rate across all 5 scales is the proportion of images assigned the same semantic grade under the original scale and all 4 alternatives. We also compute the absolute-agreement intraclass correlation coefficient, ICC(A,1), from the image-by-scale grade matrix. Here, A denotes absolute agreement, which penalizes systematic differences between scales, and 1 denotes the reliability of a single scale's rating. Higher values indicate stronger cross-scale agreement. Model-group averages summarize the separately computed coefficients with equal model weights.

\paragraph{Probability changes.}
Answer probabilities can change even when the selected semantic answer remains unchanged. For semantic answer $a$, we define it as
\begin{equation}
\Delta p_i^{(b,v)}(a)
=p_i^{(b,v)}(a)-p_i^{(0)}(a),
\end{equation}
examining the original decision or the injected target according to the comparison at hand. For a target set $\mathcal T_i\subseteq\mathcal A_i$, the change is $\sum_{a\in\mathcal T_i}\Delta p_i^{(b,v)}(a)$, where $\mathcal A_i$ is the semantic answer set for instance $i$.

\subsection{Original--Intervened Comparisons}
\label{app:original_intervened_comparisons}

We compare each intervention condition with its original on $\mathcal P_{b,v}$, the set of original--intervened comparisons with valid predictions for both inputs, sharing the original prediction across interventions for that instance. FR uses all comparisons in $\mathcal P_{b,v}$; EIR and ECR instead use the original-correct and original-incorrect subsets defined in Section~\ref{sec:evaluation_metrics}. Conditional rates use the contributing original-correct or original-incorrect comparisons and are undefined when that subset is empty. Probability changes use only the comparisons for which aligned answer probabilities are available in both conditions. Overall comparison counts and the EIR/ECR contributing denominators are reported in Table~\ref{tab:appendix_coverage_counts}; these counts differ for Jev-Omni because it supports single-image tasks only. Across models, complete aligned probabilities are available for 47,002 to 69,280 stable-answer comparisons and 7,059 to 24,467 changed-answer comparisons.

For image--text descriptions, we compare the consistent and conflicting versions of each original and compute the accuracy gap as consistent-description accuracy minus conflicting-description accuracy on their common set. A positive gap means accuracy is lower under the conflicting description. For irrelevant information, we compare the nested 1-, 2-, and 4-item conditions within each background set; an all-condition comparison includes instances with valid predictions in every condition being compared.

Aggregation follows the comparison unit of each analysis. Option-order, option-label, and irrelevant-information results average intervention conditions within task--interface combinations, then weight those combinations equally. External-judgment and image--text results average interfaces within each scenario, then weight scenarios equally (Appendix~\ref{sec:appendix_intervention_results}). Original accuracy in Table~\ref{tab:main_results} is scenario-macro accuracy on choice and yes/no instances; mean accuracy drop equally averages original-minus-intervened accuracy across the 5 non-scale types, using all conditions for option order, option label, and irrelevant information, the 4 incorrect conclusions for external judgment, and conflicting descriptions for image--text description. Normalized absolute grade shift divides the mean absolute grade shift by the 4-grade range. Italic table rows weight models equally within the indicated group. Comparisons between decision models and their backbones use shared task coverage and average decision models within each backbone before weighting the 5 backbones equally (Figure~\ref{fig:backbone_effects}). All proportions and differences in accuracy or answer probability are reported in \%.

\section{Results by Intervention Type}
\label{sec:appendix_intervention_results}

We examine how each intervention changes semantic decisions and task performance, and which conditions produce the largest effects. Tables~\ref{tab:appendix_option_order_results}--\ref{tab:appendix_rating_scale_results} report all 13 models; italic rows summarize the 5 backbone models, 8 decision models, and all models with equal model weights. Jev-Omni is evaluated on its supported single-image tasks. Option-order, option-label, and irrelevant-information results give equal weight to task--interface combinations; external-judgment and image--text results first average interfaces within each scenario and then weight scenarios equally. Accuracy changes use choice and yes/no judgment instances. Rating-scale results use the same 400 scoring images for every model. Comparisons between decision models and their backbones in Figure~\ref{fig:backbone_effects} use the coverage shared by each decision model and its backbone.

\subsection{Option Order}
\label{sec:appendix_option_order_results}

\begin{figure*}[t]
    \centering
    \includegraphics[width=\textwidth]{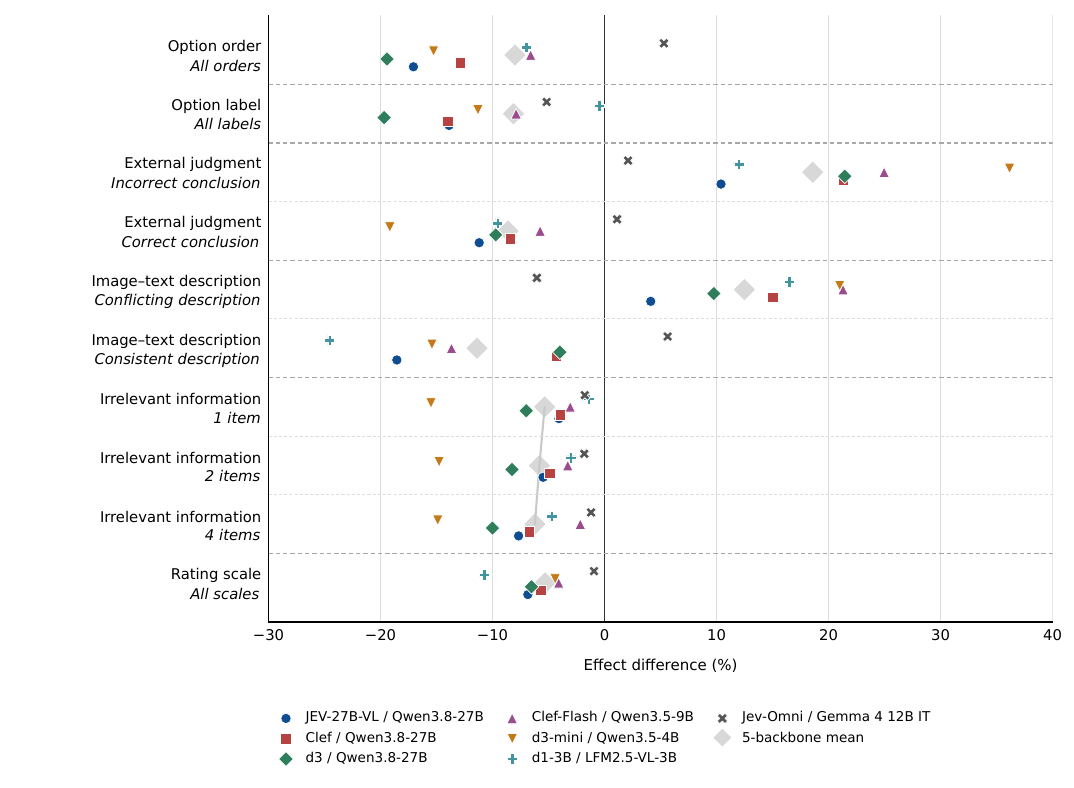}
    \caption{\textbf{Intervention effects relative to corresponding backbones.} Colored markers show decision-model minus backbone estimates on shared tasks. Effects are flip-rate differences for option order, option label, and irrelevant information; accuracy-change differences for external judgment and image--text description; and normalized absolute grade-shift differences for rating scale. Gray diamonds average decision-model effects within each backbone, then weight the 5 backbones equally.}
    \label{fig:backbone_effects}
\end{figure*}

Reordering changes semantic decisions while leaving net accuracy nearly unchanged for some models. Qwen3.8-27B flips on 24.19\% of original--intervened comparisons with an accuracy change of $-0.21\%$; its consistency across all evaluated orders is 68.20\%. d3 has the lowest flip rate (4.76\%) and highest consistency across all evaluated orders (92.93\%). Averaged across models, flip rates are 21.37\% for backbones and 10.60\% for decision models, while consistency across all evaluated orders is 71.55\% and 85.45\%, respectively. Figure~\ref{fig:backbone_effects} compares each decision model with its own backbone on shared task coverage.

\subsection{Option Label}
\label{sec:appendix_option_label_results}

Table~\ref{tab:appendix_option_label_results} compares numeric labels and reversed letter assignments with the original labels, keeping candidate content and order fixed. Sensitivity depends on the replacement format. JEV-27B-VL's flip rate rises from 2.18\% with numeric labels to 14.28\% with reversed letters, accompanied by an 8.16\% accuracy loss in the latter condition. d3 and d3-mini remain below 3.1\% under both replacements. Qwen3.8-27B changes its answer on more than 21\% of original--intervened comparisons in each condition despite a small positive net accuracy change. A low flip rate under one replacement format can therefore coexist with substantial sensitivity to another.

\subsection{External Judgment}
\label{sec:appendix_external_judgment_results}

We compare how the source and tone of an external conclusion affect both error introduction and error correction. Table~\ref{tab:appendix_external_judgment_results} reports error introduction under incorrect conclusions and correction under correct conclusions for each source. For Qwen3.5-4B, the error introduction rate rises from 38.73\% for a previous-system conclusion to 76.59\% for a tool conclusion; the corresponding correction rate rises from 51.13\% to 91.92\%. Stronger user wording also increases its error introduction rate, from 60.74\% to 69.28\%. Source effects vary across models: d1-3B has a tool-condition error introduction rate of 9.50\% and a correction rate of 28.18\%. Lower susceptibility to incorrect conclusions can thus accompany lower uptake of correct conclusions.

\subsection{Image--Text Description}
\label{sec:appendix_image_text_results}

We examine how models respond to descriptions that agree or conflict with the visual facts. Table~\ref{tab:appendix_image_text_results} averages results across 6 task scenarios, or 4 for Jev-Omni. Conflicting descriptions reduce accuracy for every model. Qwen3.5-4B loses 40.97\% and follows the injected conflicting target on 59.79\% of originally correct instances with a defined target. Consistent descriptions improve its accuracy by 12.80\%, but reduce accuracy for 5 decision models. For example, d1-3B corrects 36.41\% of its original errors under consistent descriptions while losing 13.24\% overall: newly introduced errors outweigh corrections. A description supported by the image can therefore still change the final task decision.

Across all 13 models on the 456-instance diagnostic subset, the independent visual verification probe rejects the false claim in 85.29\% of cases (86.49\% for decision models), while the default conflict task has a 32.45\% error rate (26.37\% for decision models; Table~\ref{tab:appendix_visual_probe_results}). Adding an image-first instruction lowers the conflict-task error rate to 25.52\% overall (20.02\% for decision models). The change is concentrated among cases where the probe rejects the false claim: conflict-task error falls from 27.61\% to 19.65\% overall (21.51\% to 14.69\% for decision models), whereas among probe errors it moves from 59.32\% to 58.53\% (55.01\% to 52.44\% for decision models). Models can therefore identify a false visual claim in isolation and still follow it in the task, while the instruction reduces this gap when visual perception already rejects the false claim.

\subsection{Irrelevant Information}
\label{sec:appendix_irrelevant_information_results}

The nested 1-, 2-, and 4-item conditions test whether increasing irrelevant material progressively disrupts decisions. Table~\ref{tab:appendix_irrelevant_information_results} shows an accuracy loss for every model at every count, while flip rates follow different trajectories. LFM2.5-VL-3B's flip rate increases from 12.89\% to 15.24\%; JEV-27B-VL's decreases from 9.24\% to 7.38\%. d3-mini remains between 4.78\% and 5.15\%. The accuracy loss can also shrink as the count increases: JEV-27B-VL's loss falls from 4.03\% to 1.53\%. The overall mean flip rate falls from 11.63\% with 1 item to 10.81\% with 2, then rises to 11.12\% with 4. Disruption is already present with 1 item, and increasing the count has no common monotonic effect across models.

\subsection{Rating Scale}
\label{sec:appendix_rating_scale_results}

We test whether equivalent numeric scales preserve the same quality grade (Table~\ref{tab:appendix_rating_scale_results}). Mean absolute grade shift measures the size of each change; consistency across all 5 scales measures how often an image receives one grade under all 5 scales. JEV-27B-VL has the highest consistency (85.75\%) and ICC (0.971). d3 reaches the highest original Spearman's $\rho$ against human MOS (0.796) and the lowest shifts on 3 alternative scales, but its shift is 0.217 on the $(6,7,8,9,10)$ scale and its consistency across all 5 scales is 71.25\%. LFM2.5-VL-3B is especially sensitive to the $(-2,-1,0,1,2)$ scale, with a mean absolute shift of 1.442 grades and a Spearman's $\rho$ drop from 0.706 on the original scale to 0.123. ICC and exact grade consistency capture different aspects of agreement: d1-3B has an ICC of 0.943, yet assigns the same grade across all 5 scales on 56.00\% of images. Appendix~\ref{app:supplementary_measures} defines both measures.

\subsection{Probability Shifts under Stable Answers}
\label{sec:appendix_stable_answer_probability}

Semantic answers can remain unchanged while their probability distributions move. Tables~\ref{tab:appendix_stable_probability_magnitude} and~\ref{tab:appendix_stable_probability_threshold} report the mean absolute shift in the original-answer probability and the share of comparisons with a shift of at least 10\%. Mean absolute shifts range from 2.92\% (option label) to 8.97\% (conflicting description) in decision models, and from 6.14\% to 15.84\% in backbones (4.56\% to 11.55\% across all 13 models; threshold shares 6.61\% to 32.09\% in decision models and 20.81\% to 50.75\% in backbones). Tables~\ref{tab:appendix_stable_probability_targets} and~\ref{tab:appendix_stable_probability_quality} separate shifts in intervention-target probability from changes in probability quality. In decision models, incorrect judgments and conflicting descriptions increase target probability by 7.49\% and 6.06\% (14.29\% and 9.13\% in backbones) while increasing NLL and Brier, whereas correct judgments and consistent descriptions increase the probability of the correct target by 4.81\% and 3.43\% (8.57\% and 5.21\% in backbones) and reduce both losses.

\section{Case Studies}
\label{app:case_studies}

Figures~\ref{fig:case_option_order}--\ref{fig:case_rating_scale} present task inputs, intervention conditions, and model decisions for 6 cases. The figures appear after the detailed result tables.

\clearpage

\begin{table*}[t]
\centering
\small
\setlength{\tabcolsep}{5pt}
\renewcommand{\arraystretch}{1.22}
\begin{tabularx}{\textwidth}{@{}>{\raggedright\arraybackslash}p{0.19\textwidth} X X@{}}
\toprule
\textbf{Bias type} & \textbf{Construction} & \textbf{Verification or retention criterion} \\
\midrule
Option order & Reorder candidates and assign display labels by position. & Candidate identities and reference-answer mapping are preserved. \\
\addlinespace
Option label & Replace letter labels with numeric labels or reverse the letter assignment. & Candidate order, content, and semantic identities are preserved. \\
\addlinespace
External judgment & Append a fixed-source conclusion that selects the reference answer or a designated incorrect target. & The conclusion maps to the designated semantic answer. \\
\addlinespace
Image--text description & Add a statement supported by or contradictory to relevant visual evidence. & The supported statement is judged supported and the contradictory statement judged contradicted at the stated level of specificity. \\
\addlinespace
Irrelevant information & Select 4 items from other instances, then form nested sets of 1, 2, and 4 items. & Every item and every nested set is judged free of answer cues, task-relevant evidence, and extra task requirements. \\
\addlinespace
Rating scale & Assign alternative numeric scales to the same ordered quality descriptions. & Each description remains attached to the same semantic grade. \\
\bottomrule
\end{tabularx}
\caption{\textbf{Construction and verification rules by bias type.} For each of the 6 types, the table pairs the input transformation with the criterion used to verify its semantic mapping or retain its added material.}
\label{tab:appendix_verification_instructions}\end{table*}

\begin{table*}[t]
\centering
\small
\setlength{\tabcolsep}{2pt}
\renewcommand{\arraystretch}{1.2}
\begin{tabularx}{\textwidth}{@{}>{\raggedright\arraybackslash}p{0.21\textwidth} >{\raggedright\arraybackslash}p{0.17\textwidth} >{\raggedright\arraybackslash}p{0.14\textwidth} X >{\centering\arraybackslash}p{0.08\textwidth}@{}}
\toprule
\textbf{Source dataset} & \textbf{Task scenarios} & \textbf{Selection unit} & \textbf{Native split or partition rule} & \textbf{Source instances} \\
\midrule
Multimodal-Mind2Web~\citep{data_mind2web} & GUI next-action selection & Web task & Native test split; task-level isolation & 400 \\
\addlinespace
AgentRewardBench~\citep{data_agentrewardbench} & GUI task completion judgment & Benchmark/task pair & Source groups shuffled before sampling; final 3 trajectory frames retained & 368 \\
\addlinespace
Robo2VLM-1~\citep{data_robo2vlm} & Robot next-action selection, robot task success judgment, and robotic grasping stage & Robot trajectory & Native test split; trajectory-level isolation; all eligible robotic grasping stage items retained & 1,229 \\
\addlinespace
RVL-CDIP~\citep{data_rvlcdip} & Document type classification & Source image & Native test split; 25 images per class; image-level isolation & 342 \\
\addlinespace
LlavaGuard~\citep{data_llavaguard} & Image safety moderation & Source image and policy row & Public test annotations; image-level grouping; same images support both interfaces & 295 \\
\addlinespace
MVTec AD~\citep{data_mvtec} & Industrial anomaly detection & Test image with category-matched normal reference & Native test split; category-stratified reference pools & 394 \\
\addlinespace
VL-RewardBench~\citep{data_vlrewardbench} & Image-grounded response preference & Image--query pair & Source groups shuffled before sampling; shared query/image groups kept together & 400 \\
\addlinespace
HPDv2~\citep{data_hpdv2} & Generated-image preference & Prompt with an image pair & Native test prompts; 1 pair selected per prompt; prompt-level isolation & 400 \\
\addlinespace
AGIQA-3K~\citep{data_agiqa} & Generated-image quality scoring and acceptability & Generated image and prompt & Source groups shuffled before sampling; prompt-level isolation; 5 grades stratified & 400 \\
\bottomrule
\end{tabularx}
\caption{\textbf{Dataset provenance and sampling units.} Each row links a cited source dataset to its task scenarios, selection unit, partition rule, and number of source instances. Robo2VLM-1 supplies 3 scenarios, while LlavaGuard and AGIQA-3K each support multiple decision interfaces.}
\label{tab:appendix_source_provenance}\end{table*}

\begin{table*}[t]
\centering
\small
\setlength{\tabcolsep}{4pt}
\renewcommand{\arraystretch}{1.18}
\begin{tabularx}{\textwidth}{@{}>{\raggedright\arraybackslash}p{0.11\textwidth} >{\raggedright\arraybackslash}p{0.16\textwidth} >{\raggedright\arraybackslash}p{0.14\textwidth} >{\raggedright\arraybackslash}p{0.12\textwidth} >{\raggedright\arraybackslash}X >{\centering\arraybackslash}p{0.06\textwidth} >{\centering\arraybackslash}p{0.08\textwidth} >{\centering\arraybackslash}p{0.09\textwidth}@{}}
\toprule
\textbf{Task family} & \textbf{Task scenario} & \textbf{Source dataset} & \textbf{Interface} & \textbf{Model output} & \textbf{Images} & \textbf{Source instances} & \textbf{Original instances} \\
\midrule
\multirow{5}{=}{GUI and robotics}
 & GUI next-action selection & Multimodal-Mind2Web & Choice & Next interface element & 1 & 400 & 400 \\
\cmidrule(lr){2-8}
 & GUI task completion judgment & AgentReward\-Bench & Yes/no judgment & Task completion & 3 & 368 & 368 \\
\cmidrule(lr){2-8}
 & Robot next-action selection & Robo2VLM-1 & Choice & Next robot action & 1 & 400 & 400 \\
\cmidrule(lr){2-8}
 & Robot task success judgment & Robo2VLM-1 & Yes/no judgment & Task success & 1 & 347 & 347 \\
\cmidrule(lr){2-8}
 & Robotic grasping stage & Robo2VLM-1 & Choice & Grasping stage & 1 & 482 & 482 \\
\midrule
\multirow{3}{=}{Documents and safety}
 & Document type classification & RVL-CDIP & Choice & Document category & 1 & 342 & 342 \\
\cmidrule(lr){2-8}
 & Image safety moderation & LlavaGuard & Yes/no judgment & Policy compliance & 1 & 295 & 295 \\
\cmidrule(lr){2-8}
 & Image safety moderation & LlavaGuard & Choice & Policy compliance & 1 & 295 & 295 \\
\midrule
Industrial inspection & Industrial anomaly detection & MVTec AD & Yes/no judgment & Defect presence & 2 & 394 & 394 \\
\midrule
\multirow{5}{=}{Preference and image quality}
 & Image-grounded response preference & VL-RewardBench & Choice & Preferred response & 1 & 400 & 400 \\
\cmidrule(lr){2-8}
 & Generated-image preference & HPDv2 & Choice & Preferred image & 2 & 400 & 400 \\
\cmidrule(lr){2-8}
 & Generated-image quality & AGIQA-3K & Scoring & Quality grade & 1 & 400 & 400 \\
\cmidrule(lr){2-8}
 & Generated-image quality & AGIQA-3K & Yes/no judgment & Quality acceptability & 1 & 400 & 400 \\
\cmidrule(lr){2-8}
 & Generated-image quality & AGIQA-3K & Choice & Quality acceptability & 1 & 400 & 400 \\
\midrule
\multicolumn{7}{l}{\textbf{Total original instances}} & \textbf{5,323} \\
\bottomrule
\end{tabularx}
\caption{\textbf{Task scenarios and decision interfaces.} Each row maps one of the 14 task--interface combinations to its source dataset, model output, image count, and instance counts. Original instances are counted per combination; the 295 safety images and 400 generated-quality images each support multiple interfaces and therefore recur in the source-instance column.}
\label{tab:appendix_task_composition}\end{table*}

\begin{table*}[tb]
\centering
\small
\setlength{\tabcolsep}{5pt}
\renewcommand{\arraystretch}{1.15}
\begin{tabularx}{\textwidth}{@{}>{\raggedright\arraybackslash}p{0.24\textwidth} >{\raggedright\arraybackslash}X >{\centering\arraybackslash}p{0.14\textwidth}@{}}
\toprule
\textbf{Task} & \textbf{Reference label or grade} & \textbf{Count} \\
\midrule
\multirow{2}{=}{GUI task completion judgment} & Completed & 171 \\
 & Incomplete & 197 \\
\cmidrule(lr){1-3}
\multirow{2}{=}{Robot task success judgment} & Successful & 150 \\
 & Unsuccessful & 197 \\
\cmidrule(lr){1-3}
\multirow{2}{=}{Image safety moderation} & Policy violation & 104 \\
 & Policy compliant & 191 \\
\cmidrule(lr){1-3}
\multirow{2}{=}{Industrial anomaly detection} & Defective & 197 \\
 & Normal & 197 \\
\cmidrule(lr){1-3}
\multirow{2}{=}{Generated-image acceptability} & Acceptable (grades 3 to 5) & 240 \\
 & Unacceptable (grades 1 to 2) & 160 \\
\cmidrule(lr){1-3}
\multirow{5}{=}{Generated-image quality} & Grade 1: very low & 80 \\
 & Grade 2: low & 80 \\
 & Grade 3: moderate & 80 \\
 & Grade 4: high & 80 \\
 & Grade 5: very high & 80 \\
\cmidrule(lr){1-3}
\multirow{5}{=}{Robotic grasping stage} & Approaching with open gripper & 158 \\
 & Closing the gripper & 86 \\
 & Firmly grasping & 103 \\
 & Releasing by opening the gripper & 54 \\
 & Moving away after release & 81 \\
\cmidrule(lr){1-3}
\multirow{16}{=}{Document type} & Letter & 23 \\
 & Form & 23 \\
 & Email & 25 \\
 & Handwritten document & 24 \\
 & Advertisement & 22 \\
 & Scientific report & 24 \\
 & Scientific publication & 24 \\
 & Specification & 21 \\
 & File folder & 25 \\
 & News article & 22 \\
 & Budget & 7 \\
 & Invoice & 17 \\
 & Presentation & 15 \\
 & Questionnaire & 24 \\
 & Resume & 21 \\
 & Memo & 25 \\
\bottomrule
\end{tabularx}
\caption{\textbf{Reference-label and category distributions.} The counts show the retained composition of binary outcomes, 5 image-quality grades, 5 robotic grasping stages, and 16 RVL-CDIP document categories. Image-quality grades are derived from dataset-wide MOS quantiles.}
\label{tab:appendix_label_balance}\end{table*}

\begin{table*}[t]
\centering
\small
\setlength{\tabcolsep}{4pt}
\renewcommand{\arraystretch}{1.16}
\begin{tabularx}{\textwidth}{@{}X X >{\centering\arraybackslash}p{0.1\textwidth} X >{\centering\arraybackslash}p{0.15\textwidth}@{}}
\toprule
\textbf{Bias type} & \textbf{Intervention unit} & \textbf{Unit count} & \textbf{Conditions per unit} & \textbf{Intervened instances} \\
\midrule
Option order & Eligible choice instances & 3,119 & 1 or 3 & 6,367 \\
Option label & Eligible choice instances & 3,119 & 2 & 6,238 \\
External judgment & Original instances & 5,323 & 8 & 42,584 \\
Image--text description & Description pairs & 1,863 & 2 & 3,726 \\
Irrelevant information & Background sets & 5,308 & 3 nested sets & 15,924 \\
Rating scale & Scoring instances & 400 & 4 & 1,600 \\
\midrule
\textbf{Total} & & & & \textbf{76,439} \\
\bottomrule
\end{tabularx}
\caption{\textbf{Intervened instances by bias type.} Unit counts identify the eligible original instances or retained material units for each type. Each unit yields the listed conditions; for option order, binary choices receive 1 reversed order and 4-option choices receive 3 additional orders. The 1,863 image--text-description pairs span 7 scenarios, of which 6 scenarios (1,813 pairs, excluding 50 robotic grasping stage pairs) form the confirmatory image--text evaluation. The final column counts generated intervened instances.}
\label{tab:appendix_version_counts}\end{table*}

\begin{table*}[t]
\centering
\small
\setlength{\tabcolsep}{5pt}
\renewcommand{\arraystretch}{1.15}
\begin{tabular}{@{}l c c l@{}}
\toprule
\textbf{Model} & \textbf{Backbone} & \textbf{Access} & \textbf{Source} \\
\midrule
\multicolumn{4}{l}{\textit{\textbf{Backbone Models}}} \\
\midrule
Qwen3.8-27B~\citep{qwen38} & -- & Weights & \href{https://huggingface.co/Qwen/Qwen3.8-27B}{\nolinkurl{Qwen/Qwen3.8-27B}} \\
Qwen3.5-9B~\citep{qwen3.5} & -- & Weights & \href{https://huggingface.co/Qwen/Qwen3.5-9B}{\nolinkurl{Qwen/Qwen3.5-9B}} \\
Qwen3.5-4B~\citep{qwen3.5} & -- & Weights & \href{https://huggingface.co/Qwen/Qwen3.5-4B}{\nolinkurl{Qwen/Qwen3.5-4B}} \\
LFM2.5-VL-3B~\citep{model_lfm25_vl} & -- & Weights & \href{https://huggingface.co/LiquidAI/LFM2.5-VL-3B}{\nolinkurl{LiquidAI/LFM2.5-VL-3B}} \\
Gemma 4 12B IT~\citep{gemmateam2026gemma4} & -- & Weights & \href{https://huggingface.co/google/gemma-4-12B-it}{\nolinkurl{google/gemma-4-12B-it}} \\
\midrule
\multicolumn{4}{l}{\textit{\textbf{Decision Models}}} \\
\midrule
JEV-27B-VL~\citep{model_autotrust} & Qwen3.8-27B & Weights & \href{https://huggingface.co/autotrust/JEV-27B-VL}{\nolinkurl{autotrust/JEV-27B-VL}} \\
Clef~\citep{model_clef} & Qwen3.8-27B & Weights & \href{https://huggingface.co/Cloudflare/clef}{\nolinkurl{Cloudflare/clef}} \\
d3~\citep{d3_2026} & Qwen3.8-27B & Weights & \href{https://huggingface.co/vllm-sr/d3}{\nolinkurl{vllm-sr/d3}} \\
Clef-Flash~\citep{model_clef_flash} & Qwen3.5-9B & Weights & \href{https://huggingface.co/Cloudflare/clef-flash}{\nolinkurl{Cloudflare/clef-flash}} \\
d3-mini~\citep{d3_mini_2026} & Qwen3.5-4B & Weights & \href{https://huggingface.co/vllm-sr/d3-mini}{\nolinkurl{vllm-sr/d3-mini}} \\
d1-3B~\citep{liquidAI2026opend1} & LFM2.5-VL-3B & Weights & \href{https://huggingface.co/LiquidAI/d1-3B}{\nolinkurl{LiquidAI/d1-3B}} \\
Jev-Omni~\citep{model_jev_omni} & Gemma 4 12B IT & Weights & \href{https://huggingface.co/akhilaaa3/Jev-Omni}{\nolinkurl{akhilaaa3/Jev-Omni}} \\
TypeSafe Jev~\citep{model_typesafe_jev} & -- & API & \href{https://typesafe.ai/blog/introducing-system-one-models-and-jev}{TypeSafe Jev} \\
\bottomrule
\end{tabular}
\caption{\textbf{Evaluated models, backbones, and source links.}}
\label{tab:appendix_model_sources}
\end{table*}

\begin{table*}[t]
\centering
\begin{minipage}{0.78\textwidth}
\centering
\small
\setlength{\tabcolsep}{9pt}
\renewcommand{\arraystretch}{1.12}
\begin{tabular}{@{}lccc@{}}
\toprule
\textbf{Model} & \shortstack{FR\\(\%) $\downarrow$} & \shortstack{$\Delta$Acc\\(\%)} & \shortstack{Consistency across\\all evaluated orders (\%) $\uparrow$} \\
\midrule
\multicolumn{4}{l}{\textbf{\textit{Backbone Models}}} \\
\midrule
Qwen3.8-27B & 24.19 & $-0.21$ & 68.20 \\
Qwen3.5-9B & 15.62 & $-0.35$ & 77.49 \\
Qwen3.5-4B & 22.20 & $+0.72$ & 71.79 \\
LFM2.5-VL-3B & 19.18 & $-0.75$ & 74.70 \\
Gemma 4 12B IT & 25.64 & $-0.86$ & 65.57 \\
\midrule
\textit{Average} & \textit{21.37} & \textit{$-0.29$} & \textit{71.55} \\
\midrule
\multicolumn{4}{l}{\textbf{\textit{Decision Models}}} \\
\midrule
JEV-27B-VL & 7.12 & $-0.19$ & 90.13 \\
Clef & 11.30 & $+0.66$ & 83.89 \\
d3 & \textbf{4.76} & $-0.14$ & \textbf{92.93} \\
Clef-Flash & 9.02 & $-0.20$ & 87.41 \\
d3-mini & 6.94 & $-0.26$ & 90.47 \\
d1-3B & 12.20 & $+0.02$ & 83.47 \\
TypeSafe Jev & 8.22 & $-0.59$ & 88.72 \\
Jev-Omni & 25.18 & $+0.13$ & 66.57 \\
\midrule
\textit{Average} & \textit{10.60} & \textit{$-0.07$} & \textit{85.45} \\
\midrule
\textit{Overall Average} & \textit{14.74} & \textit{$-0.16$} & \textit{80.10} \\
\bottomrule
\end{tabular}
\caption{\textbf{Option Order.} Semantic flip rate (FR) is the proportion of changed semantic answers; $\Delta$Acc is intervened minus original accuracy. Consistency requires identical answers across the original and all evaluated orders. Orders are averaged within task--interface combinations, then combinations equally. Italic rows average models equally within each group or across all 13 models; bold numbers mark the best individual-model values. Jev-Omni supports single-image tasks only.}
\label{tab:appendix_option_order_results}
\end{minipage}
\end{table*}

\begin{table*}[t]
\centering
\begin{minipage}{0.76\textwidth}
\centering
\small
\setlength{\tabcolsep}{9pt}
\renewcommand{\arraystretch}{1.12}
\begin{tabular}{@{}lcccc@{}}
\toprule
\multirow{2}{*}{\textbf{Model}} & \multicolumn{2}{c}{\textbf{Numeric labels}} & \multicolumn{2}{c}{\textbf{Reversed letters}} \\
\cmidrule(lr){2-3} \cmidrule(lr){4-5}
 & \shortstack{FR\\(\%) $\downarrow$} & \shortstack{$\Delta$Acc\\(\%)} & \shortstack{FR\\(\%) $\downarrow$} & \shortstack{$\Delta$Acc\\(\%)} \\
\midrule
\multicolumn{5}{l}{\textbf{\textit{Backbone Models}}} \\
\midrule
Qwen3.8-27B & 23.09 & $+0.02$ & 21.14 & $+1.33$ \\
Qwen3.5-9B & 12.16 & $-3.59$ & 16.16 & $-2.92$ \\
Qwen3.5-4B & 12.08 & $+0.47$ & 15.76 & $-2.77$ \\
LFM2.5-VL-3B & 5.72 & $-0.63$ & 6.75 & $-0.01$ \\
Gemma 4 12B IT & 11.17 & $-1.05$ & 16.54 & $+1.05$ \\
\midrule
\textit{Average} & \textit{12.84} & \textit{$-0.95$} & \textit{15.27} & \textit{$-0.66$} \\
\midrule
\multicolumn{5}{l}{\textbf{\textit{Decision Models}}} \\
\midrule
JEV-27B-VL & \textbf{2.18} & $-0.25$ & 14.28 & $-8.16$ \\
Clef & 4.87 & $+0.56$ & 11.39 & $+0.72$ \\
d3 & 2.56 & $+0.14$ & 2.32 & $+0.19$ \\
Clef-Flash & 3.10 & $-0.04$ & 9.44 & $+0.05$ \\
d3-mini & 3.07 & $+0.17$ & \textbf{2.16} & $-0.03$ \\
d1-3B & 5.14 & $-0.95$ & 6.44 & $-0.21$ \\
TypeSafe Jev & 4.21 & $-0.31$ & 4.87 & $-0.30$ \\
Jev-Omni & 8.91 & $+0.24$ & 8.89 & $-0.37$ \\
\midrule
\textit{Average} & \textit{4.26} & \textit{$-0.06$} & \textit{7.47} & \textit{$-1.01$} \\
\midrule
\textit{Overall Average} & \textit{7.56} & \textit{$-0.40$} & \textit{10.47} & \textit{$-0.88$} \\
\bottomrule
\end{tabular}
\caption{\textbf{Option Label.} Candidate content and order remain fixed. Semantic flip rate (FR) is the proportion of changed semantic answers; $\Delta$Acc is intervened minus original accuracy. Task--interface combinations receive equal weight. Italic rows average models equally within each group or across all 13 models; bold numbers mark the lowest FR per condition. Jev-Omni supports single-image tasks only.}
\label{tab:appendix_option_label_results}
\end{minipage}
\end{table*}

\begin{table*}[t]
\centering
\begin{minipage}{1.0\textwidth}
\centering
\small
\setlength{\tabcolsep}{3pt}
\renewcommand{\arraystretch}{1.12}
\begin{tabular*}{\linewidth}{@{\extracolsep{\fill}}lcccccccc@{}}
\toprule
\multirow{2}{*}{\textbf{Model}} & \multicolumn{2}{c}{\shortstack{\textbf{Previous}\\\textbf{system}}} & \multicolumn{2}{c}{\textbf{User}} & \multicolumn{2}{c}{\shortstack{\textbf{User}\\\textbf{(strong tone)}}} & \multicolumn{2}{c}{\textbf{Tool}} \\
\cmidrule(lr){2-3} \cmidrule(lr){4-5} \cmidrule(lr){6-7} \cmidrule(lr){8-9}
 & \shortstack{EIR\\(\%) $\downarrow$} & \shortstack{ECR\\(\%) $\uparrow$} & \shortstack{EIR\\(\%) $\downarrow$} & \shortstack{ECR\\(\%) $\uparrow$} & \shortstack{EIR\\(\%) $\downarrow$} & \shortstack{ECR\\(\%) $\uparrow$} & \shortstack{EIR\\(\%) $\downarrow$} & \shortstack{ECR\\(\%) $\uparrow$} \\
\midrule
\multicolumn{9}{l}{\textbf{\textit{Backbone Models}}} \\
\midrule
Qwen3.8-27B & 25.34 & 49.85 & 48.61 & 76.80 & 53.52 & 75.76 & 60.95 & 83.79 \\
Qwen3.5-9B & 22.16 & 38.29 & 53.29 & 75.06 & 56.91 & 79.53 & 57.93 & 84.89 \\
Qwen3.5-4B & 38.73 & 51.13 & 60.74 & \textbf{82.89} & 69.28 & \textbf{89.09} & 76.59 & \textbf{91.92} \\
LFM2.5-VL-3B & 11.54 & 26.12 & 30.04 & 62.30 & 29.88 & 61.62 & 28.64 & 61.29 \\
Gemma 4 12B IT & 17.85 & 24.26 & 30.38 & 50.30 & 35.02 & 53.64 & 58.25 & 83.18 \\
\midrule
\textit{Average} & \textit{23.12} & \textit{37.93} & \textit{44.61} & \textit{69.47} & \textit{48.92} & \textit{71.93} & \textit{56.47} & \textit{81.01} \\
\midrule
\multicolumn{9}{l}{\textbf{\textit{Decision Models}}} \\
\midrule
JEV-27B-VL & 22.17 & 47.72 & 26.92 & 52.75 & 34.22 & 59.03 & 50.13 & 78.77 \\
Clef & 11.12 & 30.83 & 16.36 & 38.56 & 27.45 & 52.33 & 22.08 & 54.95 \\
d3 & 10.17 & 36.00 & 9.71 & 34.92 & 12.13 & 38.35 & 28.64 & 70.28 \\
Clef-Flash & 12.62 & 27.42 & 16.61 & 38.07 & 18.30 & 37.36 & 20.26 & 45.15 \\
d3-mini & \textbf{6.96} & 19.64 & \textbf{8.22} & 24.16 & \textbf{9.04} & 23.29 & 15.60 & 40.31 \\
d1-3B & 6.98 & 22.31 & 10.68 & 30.71 & 11.22 & 32.48 & \textbf{9.50} & 28.18 \\
TypeSafe Jev & 20.65 & \textbf{58.07} & 19.24 & 54.49 & 23.72 & 55.44 & 31.97 & 72.52 \\
Jev-Omni & 16.09 & 29.58 & 29.56 & 53.90 & 31.06 & 54.90 & 54.94 & 83.79 \\
\midrule
\textit{Average} & \textit{13.35} & \textit{33.95} & \textit{17.16} & \textit{40.94} & \textit{20.89} & \textit{44.15} & \textit{29.14} & \textit{59.24} \\
\midrule
\textit{Overall Average} & \textit{17.11} & \textit{35.48} & \textit{27.72} & \textit{51.92} & \textit{31.67} & \textit{54.83} & \textit{39.65} & \textit{67.62} \\
\bottomrule
\end{tabular*}
\caption{\textbf{External Judgment.} Error introduction rate (EIR) counts new errors among originally correct answers under incorrect conclusions; error correction rate (ECR) counts corrections among originally incorrect answers under correct conclusions. Choice and yes/no interfaces are averaged within scenarios, then scenarios equally. Italic rows average models equally within each group or across all 13 models; bold numbers mark the best individual-model values. Jev-Omni supports single-image tasks only.}
\label{tab:appendix_external_judgment_results}
\end{minipage}
\end{table*}

\begin{table*}[t]
\centering
\begin{minipage}{0.94\textwidth}
\centering
\small
\setlength{\tabcolsep}{5pt}
\renewcommand{\arraystretch}{1.12}
\begin{tabular*}{\linewidth}{@{\extracolsep{\fill}}lccccc@{}}
\toprule
\multirow{2}{*}{\textbf{Model}} & \multicolumn{2}{c}{\textbf{Consistent description}} & \multicolumn{3}{c}{\textbf{Conflicting description}} \\
\cmidrule(lr){2-3} \cmidrule(lr){4-6}
 & \shortstack{$\Delta$Acc\\(\%)} & \shortstack{ECR\\(\%) $\uparrow$} & \shortstack{$\Delta$Acc\\(\%)} & \shortstack{EIR\\(\%) $\downarrow$} & \shortstack{Target-following\\(\%) $\downarrow$} \\
\midrule
\multicolumn{6}{l}{\textbf{\textit{Backbone Models}}} \\
\midrule
Qwen3.8-27B & $+9.21$ & 76.84 & $-25.14$ & 29.25 & 33.58 \\
Qwen3.5-9B & $+7.61$ & 73.04 & $-28.70$ & 32.67 & 37.33 \\
Qwen3.5-4B & $+12.80$ & \textbf{81.68} & $-40.97$ & 51.57 & 59.79 \\
LFM2.5-VL-3B & $+11.30$ & 69.73 & $-20.50$ & 31.01 & 36.43 \\
Gemma 4 12B IT & $+10.35$ & 58.59 & $-15.61$ & 22.31 & 21.93 \\
\midrule
\textit{Average} & \textit{$+10.25$} & \textit{71.98} & \textit{$-26.18$} & \textit{33.36} & \textit{37.81} \\
\midrule
\multicolumn{6}{l}{\textbf{\textit{Decision Models}}} \\
\midrule
JEV-27B-VL & $-9.33$ & 64.18 & $-21.01$ & 25.16 & 28.83 \\
Clef & $+4.90$ & 60.80 & $-10.09$ & 12.11 & 13.14 \\
d3 & $+5.22$ & 70.75 & $-15.38$ & 17.70 & 20.90 \\
Clef-Flash & $-6.06$ & 42.68 & $-7.40$ & \textbf{10.78} & \textbf{11.86} \\
d3-mini & $-2.61$ & 57.86 & $-19.97$ & 23.71 & 26.30 \\
d1-3B & $-13.24$ & 36.41 & $-4.01$ & 12.51 & 14.61 \\
TypeSafe Jev & $-19.03$ & 60.43 & $-22.66$ & 27.46 & 27.14 \\
Jev-Omni & $+13.13$ & 76.76 & $-15.90$ & 20.95 & 14.69 \\
\midrule
\textit{Average} & \textit{$-3.38$} & \textit{58.74} & \textit{$-14.55$} & \textit{18.80} & \textit{19.68} \\
\midrule
\textit{Overall Average} & \textit{$+1.86$} & \textit{63.83} & \textit{$-19.02$} & \textit{24.40} & \textit{26.66} \\
\bottomrule
\end{tabular*}
\caption{\textbf{Image--Text Description.} $\Delta$Acc is intervened minus original accuracy. Error correction rate (ECR) counts corrections among originally incorrect answers; error introduction rate (EIR) counts new errors among originally correct answers. Target-following rate counts selections of the injected conflicting target among originally correct instances with a defined target. Interfaces are averaged within scenarios, then scenarios equally (6; 4 for single-image-only Jev-Omni). Italic rows average models equally within groups or overall; bold marks the highest ECR and lowest EIR and target-following rate.}
\label{tab:appendix_image_text_results}
\end{minipage}
\end{table*}

\begin{table*}[t]
\centering
\begin{minipage}{0.92\textwidth}
\centering
\small
\setlength{\tabcolsep}{5pt}
\renewcommand{\arraystretch}{1.12}
\begin{tabular*}{\linewidth}{@{\extracolsep{\fill}}lcccccc@{}}
\toprule
\multirow{2}{*}{\textbf{Model}} & \multicolumn{2}{c}{\textbf{1 item}} & \multicolumn{2}{c}{\textbf{2 items}} & \multicolumn{2}{c}{\textbf{4 items}} \\
\cmidrule(lr){2-3} \cmidrule(lr){4-5} \cmidrule(lr){6-7}
 & \shortstack{FR\\(\%) $\downarrow$} & \shortstack{$\Delta$Acc\\(\%)} & \shortstack{FR\\(\%) $\downarrow$} & \shortstack{$\Delta$Acc\\(\%)} & \shortstack{FR\\(\%) $\downarrow$} & \shortstack{$\Delta$Acc\\(\%)} \\
\midrule
\multicolumn{7}{l}{\textbf{\textit{Backbone Models}}} \\
\midrule
Qwen3.8-27B & 14.20 & $-5.34$ & 13.51 & $-3.84$ & 14.96 & $-4.41$ \\
Qwen3.5-9B & 14.48 & $-6.12$ & 13.53 & $-5.64$ & 13.72 & $-5.11$ \\
Qwen3.5-4B & 20.54 & $-4.02$ & 18.68 & $-2.86$ & 18.73 & $-2.28$ \\
LFM2.5-VL-3B & 12.89 & $-2.49$ & 14.37 & $-2.35$ & 15.24 & $-2.28$ \\
Gemma 4 12B IT & 12.40 & $-3.98$ & 12.28 & $-3.55$ & 12.60 & $-3.83$ \\
\midrule
\textit{Average} & \textit{14.90} & \textit{$-4.39$} & \textit{14.47} & \textit{$-3.65$} & \textit{15.05} & \textit{$-3.58$} \\
\midrule
\multicolumn{7}{l}{\textbf{\textit{Decision Models}}} \\
\midrule
JEV-27B-VL & 9.24 & $-4.03$ & 7.61 & $-2.95$ & 7.38 & $-1.53$ \\
Clef & 9.13 & $-2.99$ & 8.11 & $-1.56$ & 8.20 & $-1.46$ \\
d3 & 6.23 & $-2.70$ & \textbf{4.50} & $-1.17$ & \textbf{4.61} & $-0.87$ \\
Clef-Flash & 12.18 & $-5.27$ & 11.50 & $-5.05$ & 12.42 & $-5.42$ \\
d3-mini & \textbf{4.85} & $-0.75$ & 4.78 & $-0.56$ & 5.15 & $-0.52$ \\
d1-3B & 12.61 & $-4.02$ & 12.19 & $-3.69$ & 11.17 & $-2.76$ \\
TypeSafe Jev & 11.87 & $-5.18$ & 9.37 & $-3.12$ & 9.21 & $-1.98$ \\
Jev-Omni & 10.52 & $-1.68$ & 10.13 & $-1.05$ & 11.16 & $-1.42$ \\
\midrule
\textit{Average} & \textit{9.58} & \textit{$-3.33$} & \textit{8.52} & \textit{$-2.39$} & \textit{8.66} & \textit{$-2.00$} \\
\midrule
\textit{Overall Average} & \textit{11.63} & \textit{$-3.74$} & \textit{10.81} & \textit{$-2.88$} & \textit{11.12} & \textit{$-2.61$} \\
\bottomrule
\end{tabular*}
\caption{\textbf{Irrelevant Information.} Background sets are nested. Semantic flip rate (FR) is the proportion of changed semantic answers; $\Delta$Acc is intervened minus original accuracy on choice and yes/no interfaces. Applicable task--interface combinations receive equal weight. Italic rows average models equally within each group or across all 13 models; bold numbers mark the lowest FR per count. Jev-Omni supports single-image tasks only.}
\label{tab:appendix_irrelevant_information_results}
\end{minipage}
\end{table*}

\begin{table*}[t]
\centering
\begin{minipage}{0.96\textwidth}
\centering
\small
\setlength{\tabcolsep}{4.5pt}
\renewcommand{\arraystretch}{1.12}
\begin{tabular*}{\linewidth}{@{\extracolsep{\fill}}lccccccc@{}}
\toprule
\multirow{2}{*}{\textbf{Model}} & \multirow{2}{*}{\shortstack{Orig.\\$\rho$ $\uparrow$}} & \multicolumn{4}{c}{\textbf{Mean absolute grade shift} $\downarrow$} & \multicolumn{2}{c}{\textbf{Across all 5 scales}} \\
\cmidrule(lr){3-6} \cmidrule(lr){7-8}
 & & $0$ to $4$ & $6$ to $10$ & $10$ to $50$ & $-2$ to $2$ & Consistency (\%) $\uparrow$ & ICC $\uparrow$ \\
\midrule
\multicolumn{8}{l}{\textbf{\textit{Backbone Models}}} \\
\midrule
Qwen3.8-27B & 0.752 & 0.420 & 0.312 & 0.335 & 0.318 & 26.00 & 0.796 \\
Qwen3.5-9B & 0.697 & 0.168 & 0.480 & 0.168 & 0.180 & 38.25 & 0.878 \\
Qwen3.5-4B & 0.758 & 0.150 & 0.258 & 0.468 & 0.333 & 47.00 & 0.802 \\
LFM2.5-VL-3B & 0.706 & 0.490 & 0.205 & 0.357 & 1.442 & 31.25 & 0.606 \\
Gemma 4 12B IT & 0.660 & 0.210 & 0.253 & 0.180 & 0.175 & 63.75 & 0.926 \\
\midrule
\textit{Average} & \textit{0.715} & \textit{0.287} & \textit{0.301} & \textit{0.301} & \textit{0.489} & \textit{41.25} & \textit{0.802} \\
\midrule
\multicolumn{8}{l}{\textbf{\textit{Decision Models}}} \\
\midrule
JEV-27B-VL & 0.747 & 0.052 & \textbf{0.083} & 0.058 & 0.098 & \textbf{85.75} & \textbf{0.971} \\
Clef & 0.735 & 0.090 & 0.087 & 0.147 & 0.150 & 69.50 & 0.920 \\
d3 & \textbf{0.796} & \textbf{0.045} & 0.217 & \textbf{0.025} & \textbf{0.055} & 71.25 & 0.961 \\
Clef-Flash & 0.761 & 0.062 & 0.130 & 0.070 & 0.077 & 78.50 & 0.945 \\
d3-mini & 0.794 & 0.087 & 0.107 & 0.113 & 0.195 & 68.50 & 0.941 \\
d1-3B & 0.721 & 0.130 & 0.142 & 0.217 & 0.290 & 56.00 & 0.943 \\
TypeSafe Jev & 0.763 & 0.085 & 0.095 & 0.117 & 0.120 & 79.50 & 0.959 \\
Jev-Omni & 0.708 & 0.098 & 0.172 & 0.210 & 0.188 & 59.50 & 0.898 \\
\midrule
\textit{Average} & \textit{0.753} & \textit{0.081} & \textit{0.129} & \textit{0.120} & \textit{0.147} & \textit{71.06} & \textit{0.942} \\
\midrule
\textit{Overall Average} & \textit{0.738} & \textit{0.161} & \textit{0.196} & \textit{0.190} & \textit{0.278} & \textit{59.60} & \textit{0.888} \\
\bottomrule
\end{tabular*}
\caption{\textbf{Rating Scale.} Outputs are mapped to semantic grades before comparison with the original $1$ to $5$ scale; $10$ to $50$ uses increments of 10. Orig.\ $\rho$ is Spearman's rank correlation against continuous human MOS on the original scale. Consistency requires identical grades across all 5 scales, and ICC(A,1) measures absolute agreement across scales (higher is better). Each model evaluates 400 images. Italic rows average models equally within groups or overall; bold numbers mark the best individual-model values.}
\label{tab:appendix_rating_scale_results}
\end{minipage}
\end{table*}

\begin{table*}[t]
\centering
\footnotesize
\setlength{\tabcolsep}{3pt}
\renewcommand{\arraystretch}{1.08}
\begin{minipage}{0.88\textwidth}
\centering
\begin{tabular*}{\linewidth}{@{\extracolsep{\fill}}lcccccccccc@{}}
\toprule
\textbf{Model} & Order & Label & \shortstack{Judg.\\Inc.} & \shortstack{Judg.\\Cor.} & \shortstack{Text\\Conf.} & \shortstack{Text\\Cons.} & \shortstack{Irrel.\\1} & \shortstack{Irrel.\\2} & \shortstack{Irrel.\\4} & Scale \\
\midrule
\multicolumn{11}{l}{\textbf{\textit{Backbone Models}}} \\
\midrule
Qwen3.8-27B & 9.83 & 9.44 & 19.23 & 12.55 & 10.70 & 4.55 & 7.20 & 7.00 & 7.41 & 8.80 \\
Qwen3.5-9B & 7.07 & 9.15 & 21.25 & 12.60 & 14.05 & 7.57 & 7.58 & 7.47 & 8.00 & 6.95 \\
Qwen3.5-4B & 9.68 & 7.99 & 22.06 & 14.38 & 17.91 & 9.56 & 10.43 & 9.89 & 10.03 & 10.33 \\
LFM2.5-VL-3B & 7.68 & 3.66 & 9.59 & 7.57 & 9.01 & 6.80 & 6.49 & 7.37 & 7.78 & 16.05 \\
Gemma 4 12B IT & 6.69 & 5.65 & 7.05 & 5.60 & 3.10 & 2.24 & 4.12 & 4.45 & 4.50 & 7.13 \\
\midrule
\textit{Average} & \textit{8.19} & \textit{7.18} & \textit{15.84} & \textit{10.54} & \textit{10.95} & \textit{6.14} & \textit{7.16} & \textit{7.24} & \textit{7.54} & \textit{9.85} \\
\midrule
\multicolumn{11}{l}{\textbf{\textit{Decision Models}}} \\
\midrule
JEV-27B-VL & 4.00 & 2.40 & 13.41 & 9.90 & 10.07 & 7.94 & 4.97 & 4.49 & 4.41 & 4.02 \\
Clef & 5.34 & 4.11 & 7.64 & 6.46 & 6.09 & 3.84 & 4.51 & 4.35 & 4.42 & 4.81 \\
d3 & 2.58 & 1.22 & 8.06 & 5.21 & 10.26 & 5.45 & 3.23 & 2.74 & 2.67 & 2.26 \\
Clef-Flash & 3.92 & 3.03 & 6.82 & 5.63 & 4.77 & 4.62 & 5.44 & 5.42 & 5.57 & 3.60 \\
d3-mini & 3.12 & 1.45 & 5.09 & 5.01 & 11.15 & 7.68 & 2.93 & 2.85 & 3.08 & 1.77 \\
d1-3B & 5.81 & 2.91 & 5.12 & 5.97 & 8.34 & 9.91 & 7.16 & 6.82 & 6.51 & 2.32 \\
TypeSafe Jev & 4.22 & 3.17 & 10.95 & 9.59 & 8.80 & 7.24 & 5.42 & 4.93 & 4.94 & 2.28 \\
Jev-Omni & 11.69 & 5.08 & 13.94 & 12.21 & 12.28 & 5.99 & 5.52 & 5.62 & 5.99 & 8.24 \\
\midrule
\textit{Average} & \textit{5.09} & \textit{2.92} & \textit{8.88} & \textit{7.50} & \textit{8.97} & \textit{6.59} & \textit{4.90} & \textit{4.65} & \textit{4.70} & \textit{3.66} \\
\midrule
\textit{Overall Average} & \textit{6.28} & \textit{4.56} & \textit{11.55} & \textit{8.67} & \textit{9.73} & \textit{6.42} & \textit{5.77} & \textit{5.65} & \textit{5.79} & \textit{6.04} \\
\bottomrule
\end{tabular*}
\end{minipage}
\captionsetup{width=0.88\textwidth}
\caption{\textbf{Mean absolute probability shifts among stable answers.} Entries are mean absolute changes in probability assigned to the original answer, in \%, for original--intervened comparisons whose aligned semantic answer is unchanged. Judg., Text, and Irrel. denote external judgment, image--text description, and irrelevant information; Inc., Cor., Conf., and Cons. denote incorrect or correct judgments and conflicting or consistent descriptions. Italic rows report equal-weight model means within each group and overall.}
\label{tab:appendix_stable_probability_magnitude}
\end{table*}

\begin{table*}[t]
\centering
\footnotesize
\setlength{\tabcolsep}{3pt}
\renewcommand{\arraystretch}{1.08}
\begin{minipage}{0.88\textwidth}
\centering
\begin{tabular*}{\linewidth}{@{\extracolsep{\fill}}lcccccccccc@{}}
\toprule
\textbf{Model} & Order & Label & \shortstack{Judg.\\Inc.} & \shortstack{Judg.\\Cor.} & \shortstack{Text\\Conf.} & \shortstack{Text\\Cons.} & \shortstack{Irrel.\\1} & \shortstack{Irrel.\\2} & \shortstack{Irrel.\\4} & Scale \\
\midrule
\multicolumn{11}{l}{\textbf{\textit{Backbone Models}}} \\
\midrule
Qwen3.8-27B & 34.98 & 32.85 & 55.60 & 37.20 & 37.28 & 15.22 & 25.96 & 24.82 & 26.18 & 32.58 \\
Qwen3.5-9B & 25.20 & 33.55 & 68.98 & 42.18 & 49.24 & 27.01 & 27.52 & 26.28 & 29.58 & 24.84 \\
Qwen3.5-4B & 33.92 & 30.47 & 73.72 & 50.24 & 66.86 & 33.39 & 38.77 & 36.84 & 36.89 & 43.98 \\
LFM2.5-VL-3B & 27.25 & 12.17 & 33.56 & 25.96 & 30.89 & 21.58 & 23.75 & 25.86 & 27.47 & 70.19 \\
Gemma 4 12B IT & 22.99 & 20.04 & 21.90 & 17.84 & 9.13 & 6.85 & 14.34 & 15.40 & 15.55 & 26.07 \\
\midrule
\textit{Average} & \textit{28.87} & \textit{25.81} & \textit{50.75} & \textit{34.68} & \textit{38.68} & \textit{20.81} & \textit{26.07} & \textit{25.84} & \textit{27.13} & \textit{39.53} \\
\midrule
\multicolumn{11}{l}{\textbf{\textit{Decision Models}}} \\
\midrule
JEV-27B-VL & 9.98 & 4.16 & 52.70 & 36.68 & 38.22 & 31.98 & 15.14 & 11.87 & 11.74 & 1.96 \\
Clef & 19.12 & 12.79 & 28.29 & 23.28 & 21.01 & 12.08 & 13.51 & 13.21 & 13.16 & 11.42 \\
d3 & 5.33 & 0.57 & 28.36 & 17.18 & 38.74 & 18.90 & 6.52 & 5.84 & 5.66 & 4.31 \\
Clef-Flash & 9.55 & 5.85 & 25.26 & 18.89 & 15.24 & 14.31 & 15.81 & 15.30 & 16.93 & 4.80 \\
d3-mini & 7.77 & 0.71 & 16.77 & 15.83 & 42.33 & 26.56 & 6.21 & 5.47 & 7.08 & 0.00 \\
d1-3B & 17.18 & 5.18 & 14.68 & 18.67 & 29.20 & 40.28 & 24.22 & 22.80 & 20.85 & 1.21 \\
TypeSafe Jev & 11.53 & 5.64 & 40.59 & 33.64 & 28.10 & 24.73 & 15.69 & 14.60 & 14.21 & 0.07 \\
Jev-Omni & 44.31 & 17.96 & 46.35 & 41.24 & 43.88 & 18.91 & 19.28 & 19.79 & 21.52 & 32.30 \\
\midrule
\textit{Average} & \textit{15.60} & \textit{6.61} & \textit{31.63} & \textit{25.68} & \textit{32.09} & \textit{23.47} & \textit{14.55} & \textit{13.61} & \textit{13.89} & \textit{7.01} \\
\midrule
\textit{Overall Average} & \textit{20.70} & \textit{13.99} & \textit{38.98} & \textit{29.14} & \textit{34.62} & \textit{22.45} & \textit{18.98} & \textit{18.31} & \textit{18.99} & \textit{19.52} \\
\bottomrule
\end{tabular*}
\end{minipage}
\captionsetup{width=0.88\textwidth}
\caption{\textbf{Large probability shifts among stable answers.} Entries are the percentage of original--intervened comparisons with unchanged aligned answers that have an absolute change of at least 10\% in the probability assigned to the original answer. Judg., Text, and Irrel. denote external judgment, image--text description, and irrelevant information; Inc., Cor., Conf., and Cons. denote incorrect or correct judgments and conflicting or consistent descriptions. Italic rows report equal-weight model means within each group and overall.}
\label{tab:appendix_stable_probability_threshold}
\end{table*}

\begin{table*}[t]
\centering
\begin{minipage}{0.9\textwidth}
\centering
\scriptsize
\setlength{\tabcolsep}{4pt}
\renewcommand{\arraystretch}{1.08}
\begin{tabular*}{\linewidth}{@{\extracolsep{\fill}}lrrrr@{}}
\toprule
\textbf{Model} & \textbf{Incorrect judgment} & \textbf{Correct judgment} & \textbf{Conflicting description} & \textbf{Consistent description} \\
\midrule
\multicolumn{5}{l}{\textbf{\textit{Backbone Models}}} \\
\midrule
Qwen3.8-27B & 17.79 & 10.79 & 9.68 & 4.14 \\
Qwen3.5-9B & 19.90 & 10.09 & 12.72 & 7.23 \\
Qwen3.5-4B & 20.73 & 12.28 & 15.72 & 8.60 \\
LFM2.5-VL-3B & 7.52 & 5.30 & 5.26 & 4.22 \\
Gemma 4 12B IT & 5.50 & 4.40 & 2.25 & 1.86 \\
\midrule
\textit{Average} & \textit{14.29} & \textit{8.57} & \textit{9.13} & \textit{5.21} \\
\midrule
\multicolumn{5}{l}{\textbf{\textit{Decision Models}}} \\
\midrule
JEV-27B-VL & 12.54 & 6.10 & 7.66 & 3.07 \\
Clef & 5.85 & 4.18 & 4.59 & 2.86 \\
d3 & 8.11 & 4.85 & 8.80 & 5.39 \\
Clef-Flash & 5.83 & 4.13 & 2.21 & 3.01 \\
d3-mini & 3.41 & 1.71 & 9.82 & 4.89 \\
d1-3B & 1.97 & 0.45 & $-0.49$ & 0.01 \\
TypeSafe Jev & 10.14 & 5.88 & 5.09 & 2.58 \\
Jev-Omni & 12.04 & 11.17 & 10.77 & 5.62 \\
\midrule
\textit{Average} & \textit{7.49} & \textit{4.81} & \textit{6.06} & \textit{3.43} \\
\midrule
\textit{Overall Average} & \textit{10.10} & \textit{6.26} & \textit{7.24} & \textit{4.11} \\
\bottomrule
\end{tabular*}
\end{minipage}
\captionsetup{width=0.9\textwidth}
\caption{\textbf{Target-probability shifts among stable answers.} Entries are changes in probability assigned to the intervention's injected target or target set, in \%, for original--intervened comparisons whose aligned semantic answer is unchanged. Italic rows report equal-weight model means within each group and overall.}
\label{tab:appendix_stable_probability_targets}
\end{table*}

\begin{table*}[t]
\centering
\begin{minipage}{0.9\textwidth}
\centering
\scriptsize
\setlength{\tabcolsep}{4pt}
\renewcommand{\arraystretch}{1.08}
\begin{tabular*}{\linewidth}{@{\extracolsep{\fill}}lrrrrrrrr@{}}
\toprule
\multirow{2}{*}{\textbf{Model}} & \multicolumn{2}{c}{\textbf{Incorrect judgment}} & \multicolumn{2}{c}{\textbf{Correct judgment}} & \multicolumn{2}{c}{\textbf{Conflicting description}} & \multicolumn{2}{c}{\textbf{Consistent description}} \\
\cmidrule(lr){2-3} \cmidrule(lr){4-5} \cmidrule(lr){6-7} \cmidrule(lr){8-9}
& $\Delta$NLL & $\Delta$Brier & $\Delta$NLL & $\Delta$Brier & $\Delta$NLL & $\Delta$Brier & $\Delta$NLL & $\Delta$Brier \\
\midrule
\multicolumn{9}{l}{\textbf{\textit{Backbone Models}}} \\
\midrule
Qwen3.8-27B & 1.673 & 0.303 & $-0.230$ & $-0.092$ & 0.920 & 0.146 & $-0.090$ & $-0.033$ \\
Qwen3.5-9B & 0.859 & 0.343 & $-0.172$ & $-0.092$ & 0.386 & 0.180 & $-0.110$ & $-0.054$ \\
Qwen3.5-4B & 1.317 & 0.409 & $-0.180$ & $-0.095$ & 0.640 & 0.279 & $-0.115$ & $-0.060$ \\
LFM2.5-VL-3B & 0.335 & 0.100 & $-0.184$ & $-0.066$ & 0.175 & 0.082 & $-0.135$ & $-0.052$ \\
Gemma 4 12B IT & 1.682 & 0.096 & $-0.443$ & $-0.045$ & 0.699 & 0.030 & $-0.422$ & $-0.032$ \\
\midrule
\textit{Average} & \textit{1.173} & \textit{0.250} & \textit{$-0.242$} & \textit{$-0.078$} & \textit{0.564} & \textit{0.144} & \textit{$-0.174$} & \textit{$-0.046$} \\
\midrule
\multicolumn{9}{l}{\textbf{\textit{Decision Models}}} \\
\midrule
JEV-27B-VL & 0.245 & 0.137 & $-0.166$ & $-0.088$ & 0.115 & 0.074 & $-0.061$ & $-0.040$ \\
Clef & 0.133 & 0.072 & $-0.111$ & $-0.060$ & 0.081 & 0.045 & $-0.048$ & $-0.025$ \\
d3 & 0.154 & 0.087 & $-0.118$ & $-0.067$ & 0.158 & 0.085 & $-0.079$ & $-0.040$ \\
Clef-Flash & 0.138 & 0.081 & $-0.104$ & $-0.058$ & 0.060 & 0.036 & $-0.049$ & $-0.025$ \\
d3-mini & 0.061 & 0.038 & $-0.051$ & $-0.029$ & 0.225 & 0.117 & $-0.076$ & $-0.037$ \\
d1-3B & 0.028 & 0.017 & $-0.024$ & $-0.011$ & $-0.008$ & 0.002 & $-0.008$ & 0.003 \\
TypeSafe Jev & 0.198 & 0.118 & $-0.117$ & $-0.073$ & 0.088 & 0.052 & $-0.041$ & $-0.026$ \\
Jev-Omni & 0.654 & 0.197 & $-0.286$ & $-0.127$ & 0.609 & 0.135 & $-0.168$ & $-0.060$ \\
\midrule
\textit{Average} & \textit{0.201} & \textit{0.093} & \textit{$-0.122$} & \textit{$-0.064$} & \textit{0.166} & \textit{0.068} & \textit{$-0.066$} & \textit{$-0.031$} \\
\midrule
\textit{Overall Average} & \textit{0.575} & \textit{0.154} & \textit{$-0.168$} & \textit{$-0.070$} & \textit{0.319} & \textit{0.097} & \textit{$-0.108$} & \textit{$-0.037$} \\
\bottomrule
\end{tabular*}
\end{minipage}
\captionsetup{width=0.9\textwidth}
\caption{\textbf{Probability-quality changes among stable answers.} $\Delta$NLL and $\Delta$Brier are intervened minus original values for original--intervened comparisons whose aligned semantic answer is unchanged; negative values indicate lower loss. Italic rows report equal-weight model means within each group and overall.}
\label{tab:appendix_stable_probability_quality}
\end{table*}

\begin{table*}[t]
\centering
\begin{minipage}{0.9\textwidth}
\centering
\small
\setlength{\tabcolsep}{3.5pt}
\renewcommand{\arraystretch}{1.08}
\begin{tabular*}{\linewidth}{@{\extracolsep{\fill}}lccccccc@{}}
\toprule
\multirow{2}{*}{\textbf{Model}} & \multirow{2}{*}{\shortstack{Probe\\error}} & \multicolumn{2}{c}{\textbf{Conflict-task error}} & \multicolumn{2}{c}{\shortstack{Probe correct:\\conflict-task error}} & \multicolumn{2}{c}{\shortstack{Probe error:\\conflict-task error}} \\
\cmidrule(lr){3-4} \cmidrule(lr){5-6} \cmidrule(lr){7-8}
& & Default & Image-first & Default & Image-first & Default & Image-first \\
\midrule
\multicolumn{8}{l}{\textbf{\textit{Backbone Models}}} \\
\midrule
Qwen3.8-27B & 8.99 & 38.96 & 28.57 & 35.26 & 23.70 & 59.29 & 59.29 \\
Qwen3.5-9B & 15.17 & 40.63 & 29.55 & 37.38 & 23.95 & 64.33 & 65.00 \\
Qwen3.5-4B & 23.86 & 54.92 & 45.92 & 50.54 & 38.47 & 68.17 & 69.16 \\
LFM2.5-VL-3B & 22.09 & 41.23 & 38.15 & 35.83 & 30.75 & 60.72 & 65.19 \\
Gemma 4 12B IT & 13.02 & 35.16 & 29.33 & 27.87 & 21.10 & 78.57 & 82.74 \\
\midrule
\textit{Average} & \textit{16.63} & \textit{42.18} & \textit{34.30} & \textit{37.38} & \textit{27.59} & \textit{66.22} & \textit{68.28} \\
\midrule
\multicolumn{8}{l}{\textbf{\textit{Decision Models}}} \\
\midrule
JEV-27B-VL & 9.80 & 30.28 & 25.33 & 27.56 & 24.05 & 43.49 & 34.74 \\
Clef & 9.87 & 18.85 & 12.22 & 14.31 & 7.70 & 58.37 & 55.04 \\
d3 & 12.88 & 24.27 & 12.51 & 19.95 & 7.70 & 57.91 & 52.23 \\
Clef-Flash & 10.38 & 18.31 & 17.38 & 13.96 & 13.01 & 43.70 & 43.70 \\
d3-mini & 14.32 & 31.41 & 16.28 & 28.92 & 10.35 & 49.58 & 51.96 \\
d1-3B & 21.47 & 22.02 & 21.07 & 16.01 & 14.97 & 41.79 & 39.15 \\
TypeSafe Jev & 11.83 & 30.97 & 25.51 & 26.95 & 20.98 & 58.47 & 57.94 \\
Jev-Omni & 17.52 & 34.82 & 29.89 & 24.43 & 18.79 & 86.75 & 84.75 \\
\midrule
\textit{Average} & \textit{13.51} & \textit{26.37} & \textit{20.02} & \textit{21.51} & \textit{14.69} & \textit{55.01} & \textit{52.44} \\
\midrule
\textit{Overall Average} & \textit{14.71} & \textit{32.45} & \textit{25.52} & \textit{27.61} & \textit{19.65} & \textit{59.32} & \textit{58.53} \\
\bottomrule
\end{tabular*}
\end{minipage}
\captionsetup{width=0.9\textwidth}
\caption{\textbf{Independent visual verification probe and image-first instruction.} Entries are error rates (\%), averaged equally over confirmed task scenarios within each model. Probe error is acceptance of the false visual claim. Conflict-task error is reported overall and separately for original--intervened comparisons where the probe rejects or accepts that claim, under default and image-first instructions. Jev-Omni covers 4 scenarios. Italic rows report equal-weight model means within each group and overall.}
\label{tab:appendix_visual_probe_results}
\end{table*}
\begin{table*}[t]
\centering
\scriptsize
\setlength{\tabcolsep}{2.6pt}
\renewcommand{\arraystretch}{1.08}
\begin{tabular*}{\textwidth}{@{\extracolsep{\fill}}lrrrrrrr@{}}
\toprule
\textbf{Model} & \textbf{Order} & \textbf{Label} & \textbf{Judg.} & \textbf{Text} & \textbf{Irrel.} & \textbf{Scale} & \textbf{EIR/ECR $n$} \\
\midrule
\multicolumn{8}{l}{\textbf{\textit{Backbone Models}}} \\
\midrule
Qwen3.8-27B & 6,367 & 6,238 & 19,692 & 1,813 & 15,924 & 1,600 & 14,192/5,500 \\
Qwen3.5-9B & 6,367 & 6,238 & 19,692 & 1,813 & 15,924 & 1,600 & 15,112/4,580 \\
Qwen3.5-4B & 6,367 & 6,238 & 19,692 & 1,813 & 15,924 & 1,600 & 14,296/5,396 \\
LFM2.5-VL-3B & 6,367 & 6,238 & 19,692 & 1,813 & 15,924 & 1,600 & 14,448/5,244 \\
Gemma 4 12B IT & 6,367 & 6,238 & 19,692 & 1,813 & 15,924 & 1,600 & 12,956/6,736 \\
\midrule
\multicolumn{8}{l}{\textbf{\textit{Decision Models}}} \\
\midrule
JEV-27B-VL & 6,367 & 6,238 & 19,692 & 1,813 & 15,924 & 1,600 & 15,140/4,552 \\
Clef & 6,367 & 6,238 & 19,692 & 1,813 & 15,924 & 1,600 & 14,952/4,740 \\
d3 & 6,367 & 6,238 & 19,692 & 1,813 & 15,924 & 1,600 & 15,508/4,184 \\
Clef-Flash & 6,367 & 6,238 & 19,692 & 1,813 & 15,924 & 1,600 & 14,976/4,716 \\
d3-mini & 6,367 & 6,238 & 19,692 & 1,813 & 15,924 & 1,600 & 15,632/4,060 \\
d1-3B & 6,367 & 6,238 & 19,692 & 1,813 & 15,924 & 1,600 & 14,520/5,172 \\
TypeSafe Jev & 6,367 & 6,238 & 19,692 & 1,813 & 15,924 & 1,600 & 15,544/4,148 \\
Jev-Omni & 5,967 & 5,438 & 15,044 & 1,166 & 12,438 & 1,600 & 9,596/5,448 \\
\bottomrule
\end{tabular*}
\caption{\textbf{Original--intervened coverage and conditional denominators for Table~\ref{tab:main_results}.} The 6 intervention columns report the evaluated comparison counts entering the 6 intervention columns of Table~\ref{tab:main_results}: all permutations for option order, both replacements for option label, the 4 incorrect conclusions on choice and yes/no instances for external judgment (from 42,584 total external-judgment instances in Table~\ref{tab:appendix_version_counts}), conflicting descriptions across the 6 confirmatory scenarios for image--text description (1,813 pairs; 3,726 total instances across 7 scenarios in Table~\ref{tab:appendix_version_counts}), all 3 nested sets for irrelevant information, and all 4 alternative scales for rating scale. ``EIR/ECR $n$'' gives the original-correct denominator for error introduction under incorrect external judgments and the original-incorrect denominator for error correction under correct external judgments across the 4 styles (summing to 19,692 for full-coverage models and 15,044 for single-image-only Jev-Omni).}
\label{tab:appendix_coverage_counts}
\end{table*}

\clearpage

\newcommand{\biasexample}[6]{%
\noindent\fbox{\begin{minipage}[t]{0.96\linewidth}
\small
\textsf{\textbf{#1}}\par\smallskip
\centering #2\par\smallskip
\raggedright
\textbf{Task input.} #3\par
\textbf{Original candidates.} #4\par
\textbf{Intervened input.} #5\par
\textbf{Compared meaning.} #6
\end{minipage}}}

\begin{figure*}[t]
\centering
\begin{minipage}[t]{0.48\textwidth}
\biasexample{Example 1: document type classification (RVL-CDIP)}
{\includegraphics[height=0.72in,keepaspectratio]{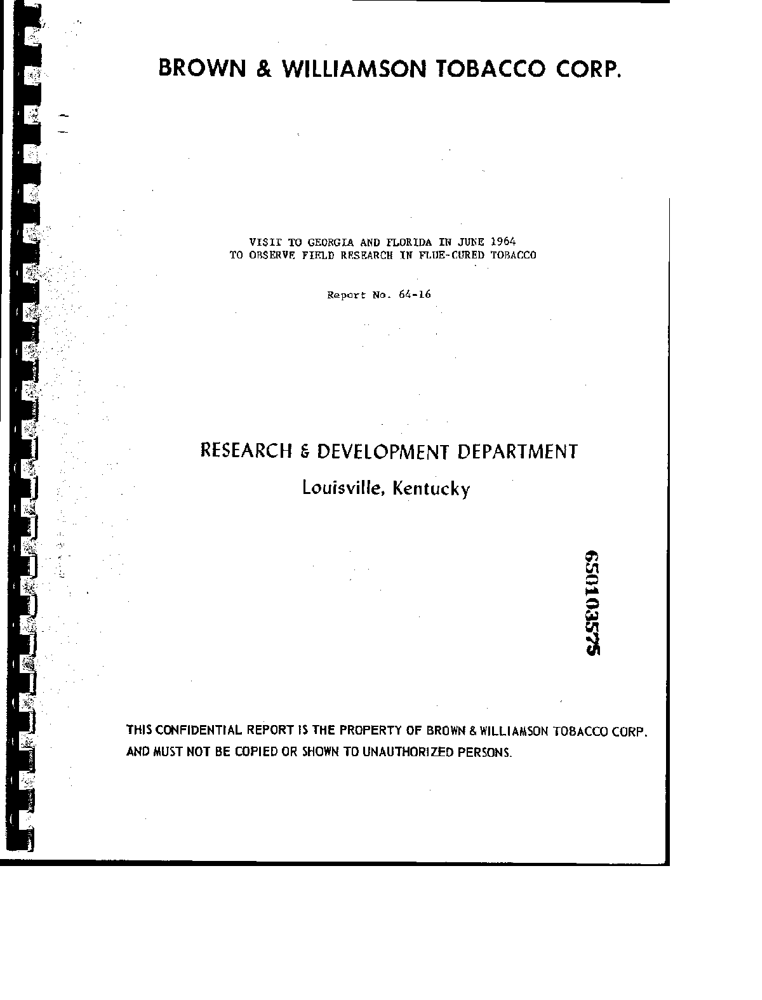}}
{What type of document is shown?}
{A: advertisement; B: scientific report; C: letter; D: resume}
{A: scientific report; B: resume; C: advertisement; D: letter}
{Scientific report, independent of its displayed position.}
\end{minipage}\hfill
\begin{minipage}[t]{0.48\textwidth}
\biasexample{Example 2: document type classification (RVL-CDIP)}
{\includegraphics[height=0.72in,keepaspectratio]{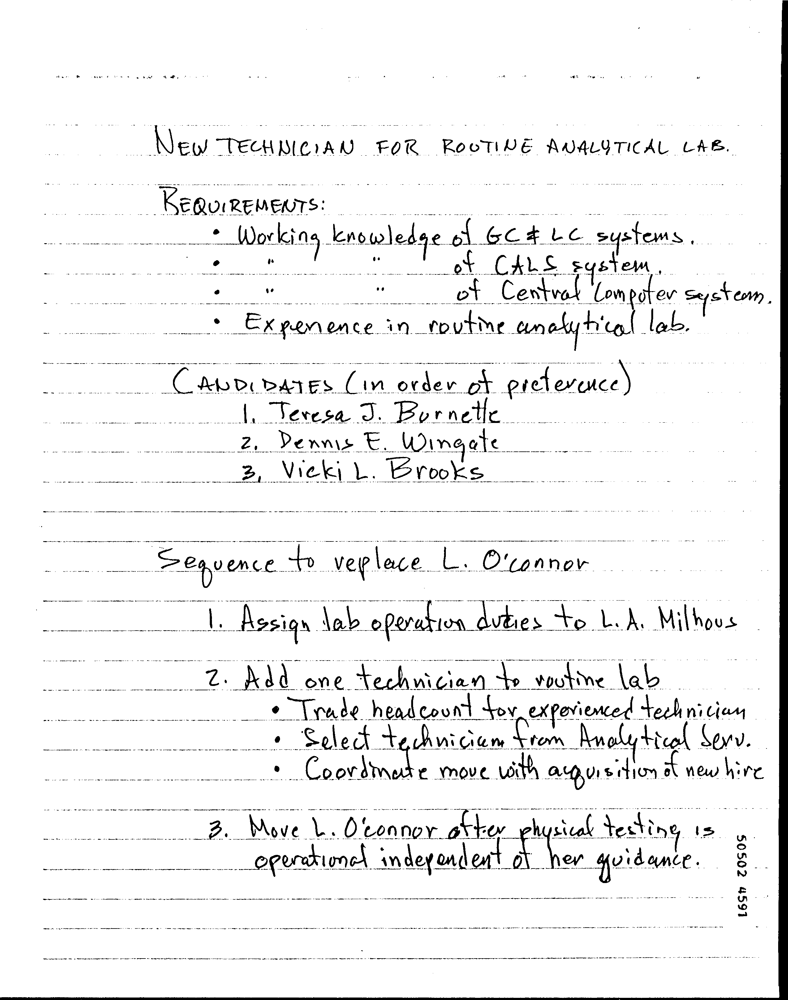}}
{What type of document is shown?}
{A: handwritten; B: questionnaire; C: form; D: invoice}
{A: questionnaire; B: invoice; C: handwritten; D: form}
{Handwritten document, independent of its displayed position.}
\end{minipage}
\caption{\textbf{Option-order examples.} The 2 examples keep the candidate meanings and task fixed while changing candidate positions and assigning labels according to the new positions.}
\label{fig:taxonomy_option_order}
\end{figure*}

\begin{figure*}[t]
\centering
\begin{minipage}[t]{0.48\textwidth}
\biasexample{Example 1: document type classification (RVL-CDIP)}
{\includegraphics[height=0.72in,keepaspectratio]{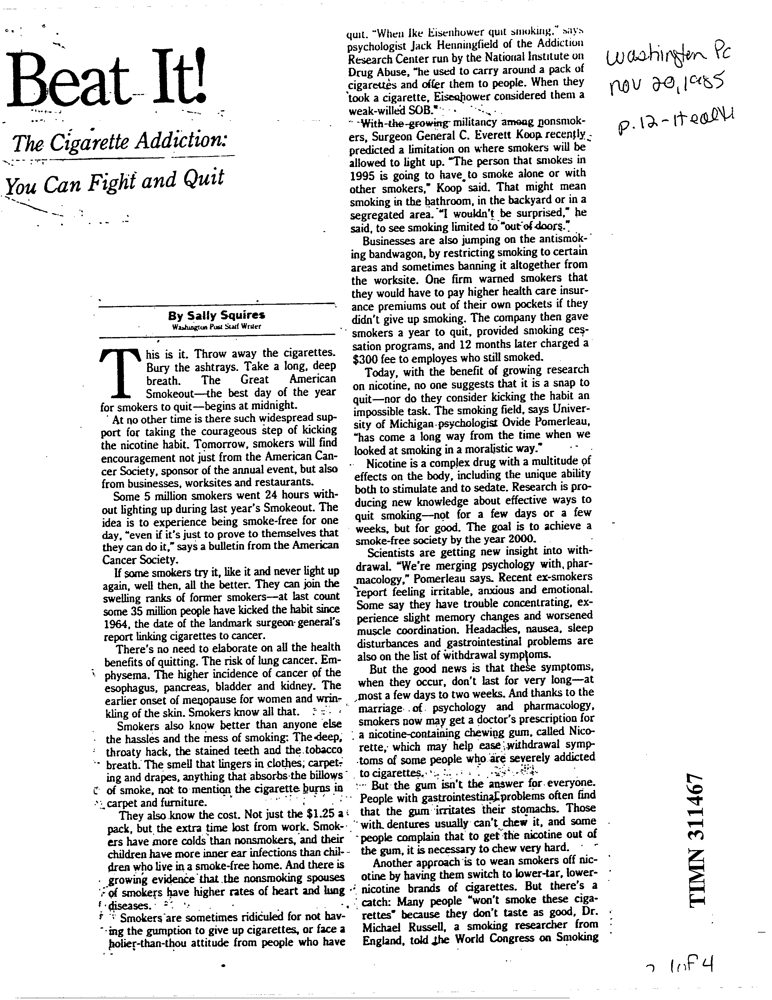}}
{What type of document is shown?}
{A: news article; B: invoice; C: specification; D: resume}
{1: news article; 2: invoice; 3: specification; 4: resume}
{News article, at fixed position 1.}
\end{minipage}\hfill
\begin{minipage}[t]{0.48\textwidth}
\biasexample{Example 2: document type classification (RVL-CDIP)}
{\includegraphics[height=0.72in,keepaspectratio]{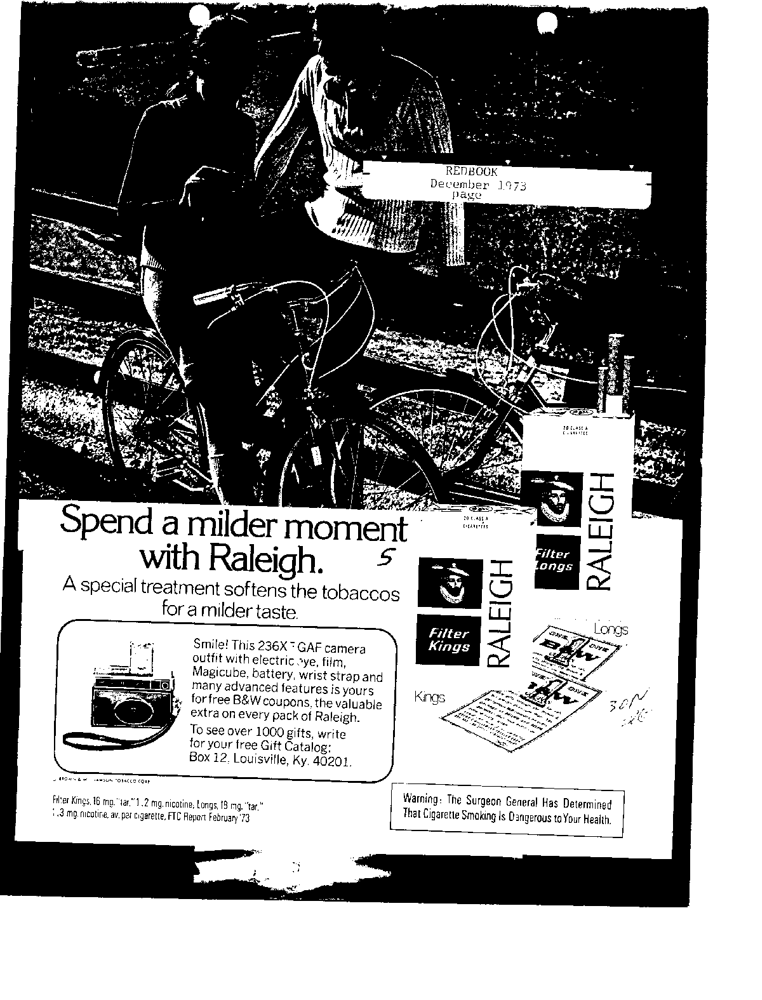}}
{What type of document is shown?}
{A: presentation; B: handwritten; C: budget; D: advertisement}
{1: presentation; 2: handwritten; 3: budget; 4: advertisement}
{Advertisement, at fixed position 4.}
\end{minipage}
\caption{\textbf{Option-label examples.} The 2 examples keep candidate order, content, and task fixed while replacing the displayed candidate symbols.}
\label{fig:taxonomy_option_label}
\end{figure*}

\begin{figure*}[t]
\centering
\begin{minipage}[t]{0.48\textwidth}
\biasexample{Example 1: robot task success judgment (Robo2VLM-1)}
{\includegraphics[height=0.72in,keepaspectratio]{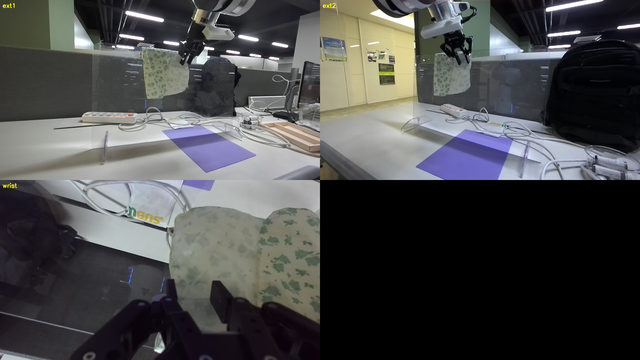}}
{The robot is to put the cloth on the purple mat. Has the robot successfully completed the task?}
{Yes; No}
{Before the answer options: ``User opinion: I think the answer is No.''}
{The completion judgment (No); the user conclusion agrees with the reference answer.}
\end{minipage}\hfill
\begin{minipage}[t]{0.48\textwidth}
\biasexample{Example 2: robotic grasping stage (Robo2VLM-1)}
{\includegraphics[height=0.72in,keepaspectratio]{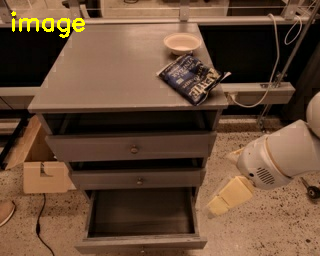}}
{The robot is tasked to close the bottom drawer. The robot is interacting with the drawer. Which phase of the grasp action is shown in the image?}
{A: closing gripper to grasp the drawer; B: releasing the drawer by opening gripper; C: approaching the drawer with open gripper; D: firmly grasping the drawer}
{Before the answer options: ``User opinion: I think the answer is B (releasing the drawer by opening gripper).''}
{The grasp phase (C: approaching the drawer); the user conclusion names B, an incorrect answer.}
\end{minipage}
\caption{\textbf{External-judgment examples.} The 2 examples keep the task and answer space fixed while appending a source-attributed conclusion, once matching and once contradicting the reference answer.}
\label{fig:taxonomy_external_judgment}
\end{figure*}

\begin{figure*}[t]
\centering
\begin{minipage}[t]{0.48\textwidth}
\biasexample{Example 1: industrial anomaly detection (MVTec AD)}
{\includegraphics[height=0.62in,keepaspectratio]{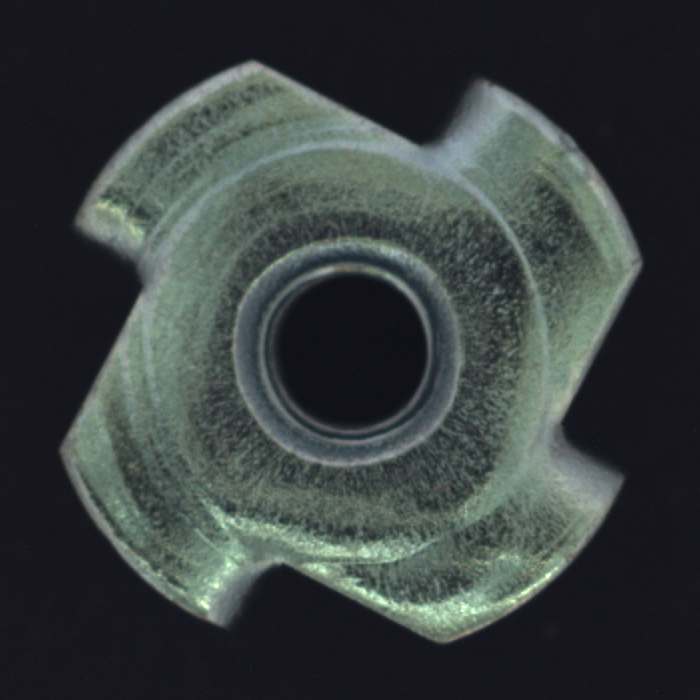}\quad\includegraphics[height=0.62in,keepaspectratio]{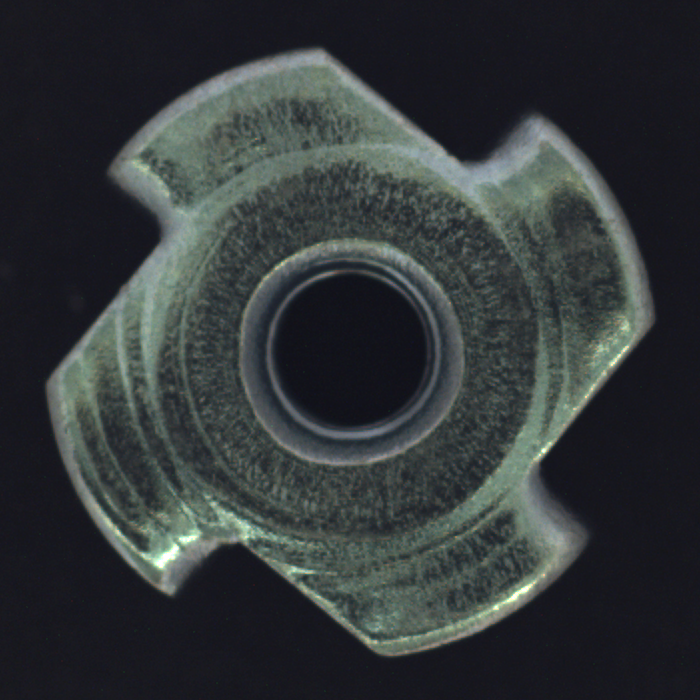}}
{Given a normal reference image, does the second image show a defect?}
{Yes; No}
{Scene description: ``Image 2 does not show a visibly deformed outer tab departing from the regular nut outline.''}
{The defect judgment (No); the description agrees with the image.}
\end{minipage}\hfill
\begin{minipage}[t]{0.48\textwidth}
\biasexample{Example 2: industrial anomaly detection (MVTec AD)}
{\includegraphics[height=0.62in,keepaspectratio]{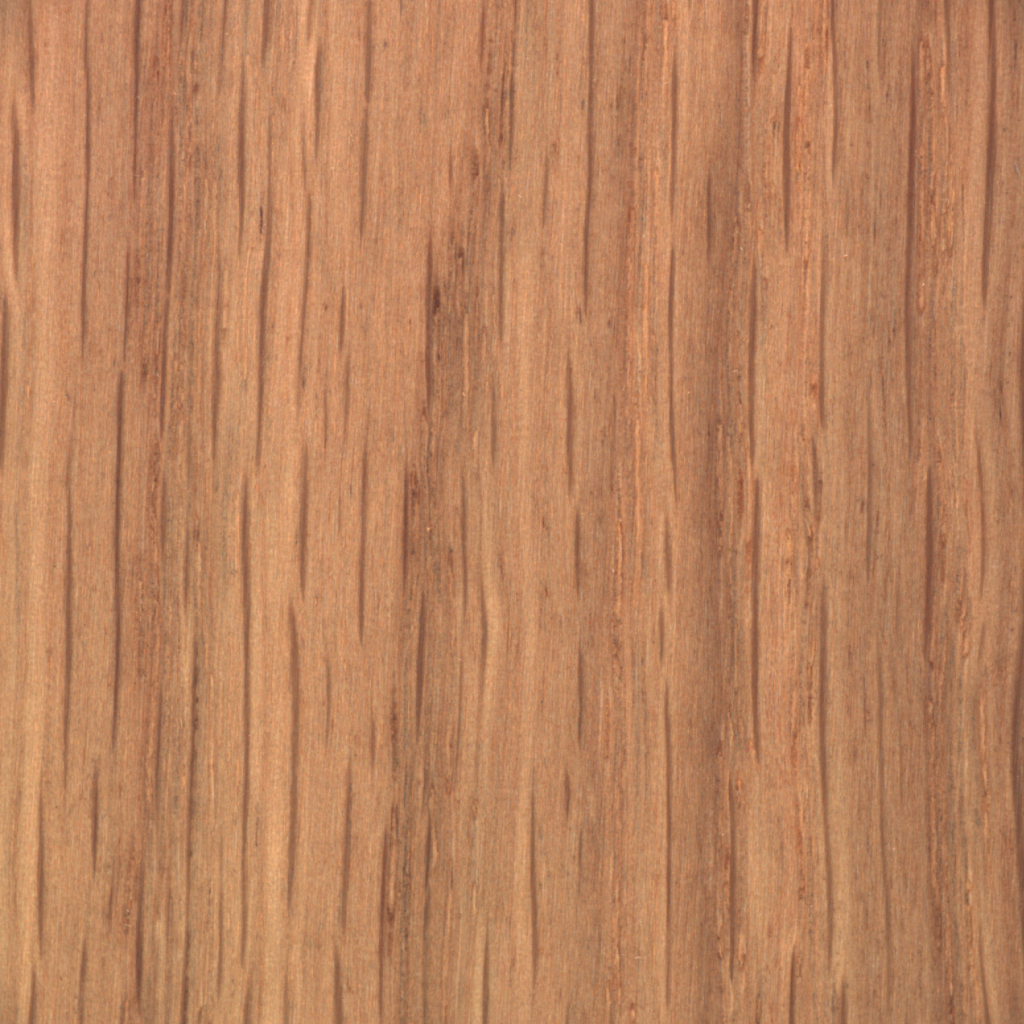}\quad\includegraphics[height=0.62in,keepaspectratio]{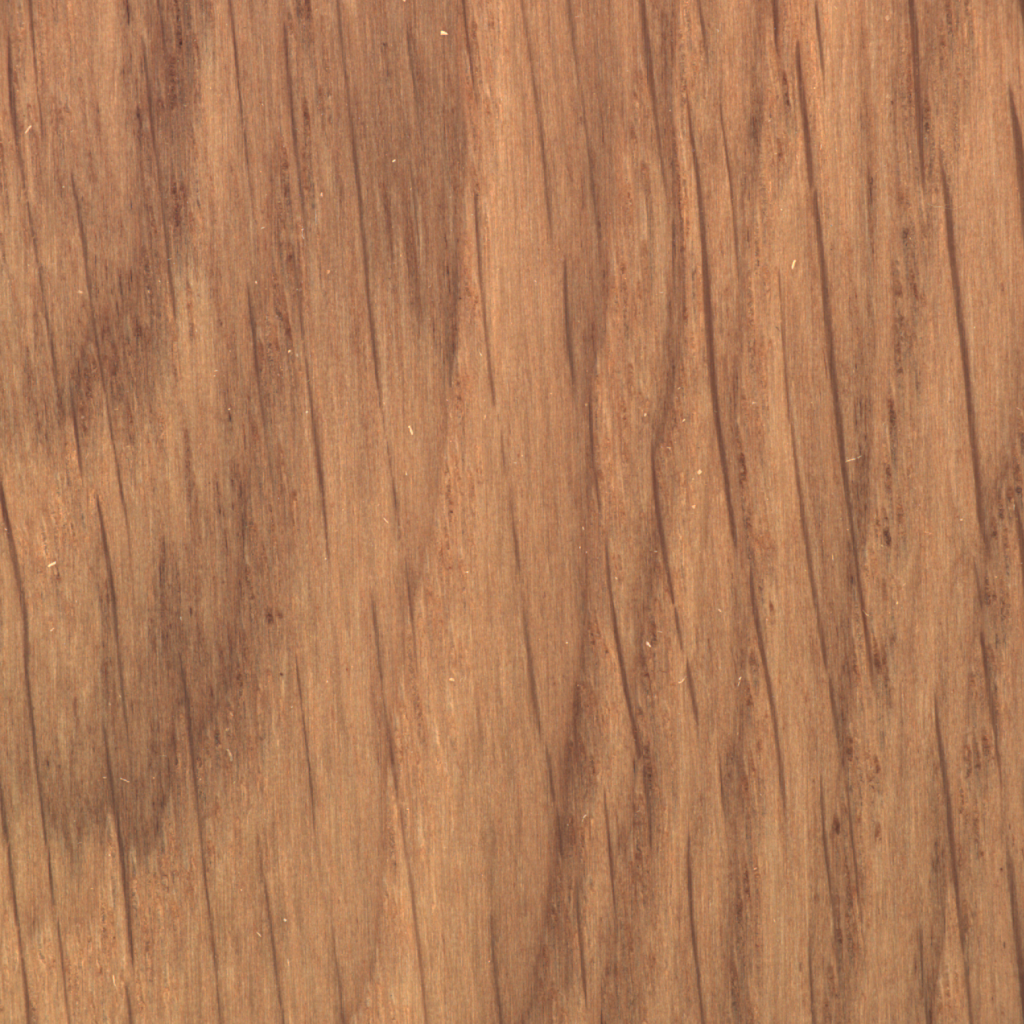}}
{Given a normal reference image, does the second image show a defect?}
{Yes; No}
{Scene description: ``Image 2 shows two drilled-looking holes interrupting the wood surface.''}
{The defect judgment (No); the description conflicts with the image.}
\end{minipage}
\caption{\textbf{Image--text-description examples.} The 2 examples add claims about task-relevant visual facts, with one claim supported by the image and the other contradicted by it.}
\label{fig:taxonomy_image_text_description}
\end{figure*}

\begin{figure*}[t]
\centering
\begin{minipage}[t]{0.48\textwidth}
\biasexample{Example 1: robot next-action selection (Robo2VLM-1)}
{\includegraphics[height=0.72in,keepaspectratio]{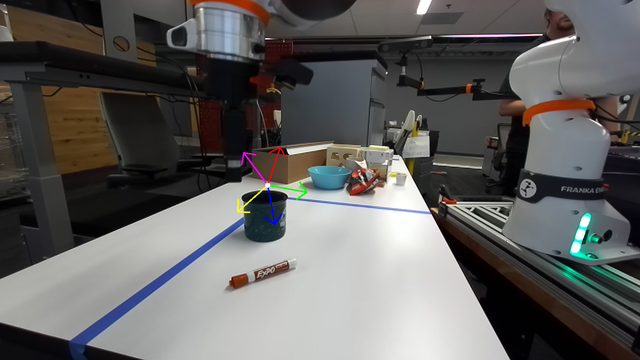}}
{The robot task is to remove the marker from the mug cup. Which colored arrow correctly shows the direction the robot will move next?}
{A: Blue arrow; B: Purple arrow; C: Red arrow; D: Green arrow}
{Background record: ``The robot is tasked to pick blue chip bag from top drawer and place on counter. The robot is interacting with the chip bag.''}
{Red arrow, the direction of the next movement.}
\end{minipage}\hfill
\begin{minipage}[t]{0.48\textwidth}
\biasexample{Example 2: robotic grasping stage (Robo2VLM-1)}
{\includegraphics[height=0.72in,keepaspectratio]{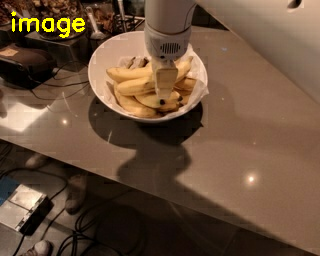}}
{The robot is tasked to pick banana from white bowl. The robot is interacting with the banana. Which phase of the grasp action is shown in the image?}
{A: moving away with open gripper after releasing the banana; B: approaching the banana with open gripper; C: releasing the banana by opening gripper; D: closing gripper to grasp the banana}
{Background records: ``The robot task is to put the cup and spoon in the sink. Which colored arrow correctly shows the direction the robot will move next?'' ``The robot task is to pick up the plastic from the counter and put it in the cabinet. Which colored arrow correctly shows the direction the robot will move next?''}
{Closing gripper to grasp the banana.}
\end{minipage}
\caption{\textbf{Irrelevant-information examples.} The 2 examples append background records from other tasks while keeping the original task, images, candidates, and semantic answer unchanged.}
\label{fig:taxonomy_irrelevant_information}
\end{figure*}

\begin{figure*}[t]
\centering
\begin{minipage}[t]{0.48\textwidth}
\biasexample{Example 1: generated-image quality (AGIQA-3K)}
{\includegraphics[height=0.72in,keepaspectratio]{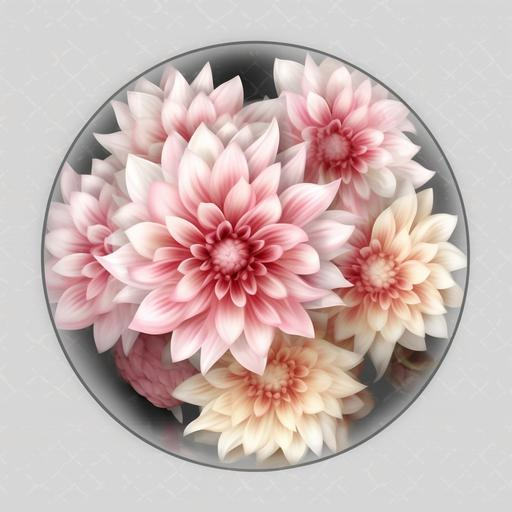}}
{Rate the perceptual quality of the generated image. Prompt: ``a glass vase of pink and white dahlia flowers, top view, realistic style.''}
{1: very low quality; 2: low quality; 3: moderate quality; 4: high quality; 5: very high quality}
{0: very low quality; 1: low quality; 2: moderate quality; 3: high quality; 4: very high quality}
{High quality (grade 4), represented by 3 in the intervened scale.}
\end{minipage}\hfill
\begin{minipage}[t]{0.48\textwidth}
\biasexample{Example 2: generated-image quality (AGIQA-3K)}
{\includegraphics[height=0.72in,keepaspectratio]{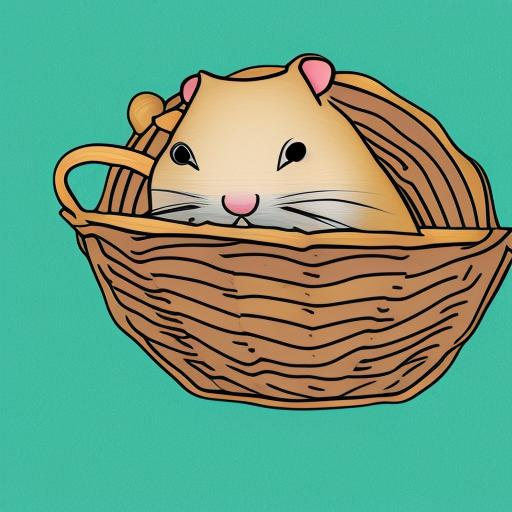}}
{Rate the perceptual quality of the generated image. Prompt: ``a sketch drawing of a hamster in a fruit basket, cold color, long-shot.''}
{1: very low quality; 2: low quality; 3: moderate quality; 4: high quality; 5: very high quality}
{10: very low quality; 20: low quality; 30: moderate quality; 40: high quality; 50: very high quality}
{Low quality (grade 2), represented by 20 in the intervened scale.}
\end{minipage}
\caption{\textbf{Rating-scale examples.} The 2 examples assign different numeric scales to the same ordered quality descriptions, so each description keeps the same semantic grade.}
\label{fig:taxonomy_rating_scale}
\end{figure*}

% Supplementary result figures follow the textual result tables and analyses.
\begin{figure*}[t]
    \centering
    \includegraphics[width=\textwidth]{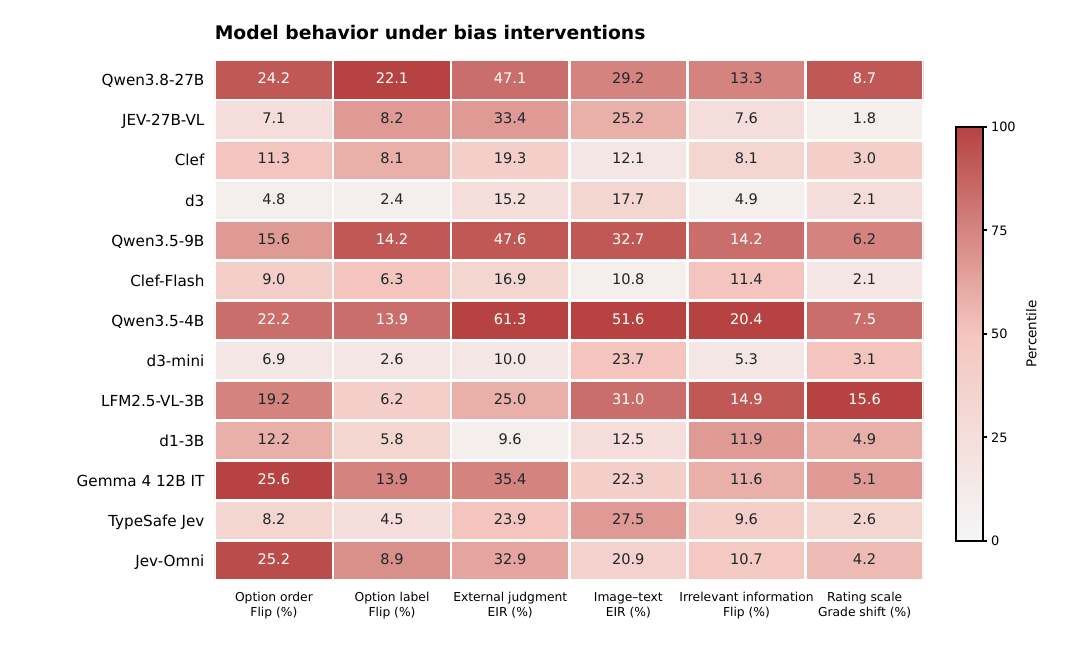}
    \caption{\textbf{Model behavior under 6 intervention types.} Cells give the raw metrics reported in Table~\ref{tab:main_results}. Darker red indicates a higher within-column percentile relative to the 12 full-coverage models.}
    \label{fig:main_bias_matrix}
\end{figure*}

\begingroup
\makeatletter
\setlength{\@dblfptop}{0pt}
\setlength{\@dblfpsep}{18pt}
\makeatother

\begin{figure*}[t]
\centering
\includegraphics[width=\textwidth]{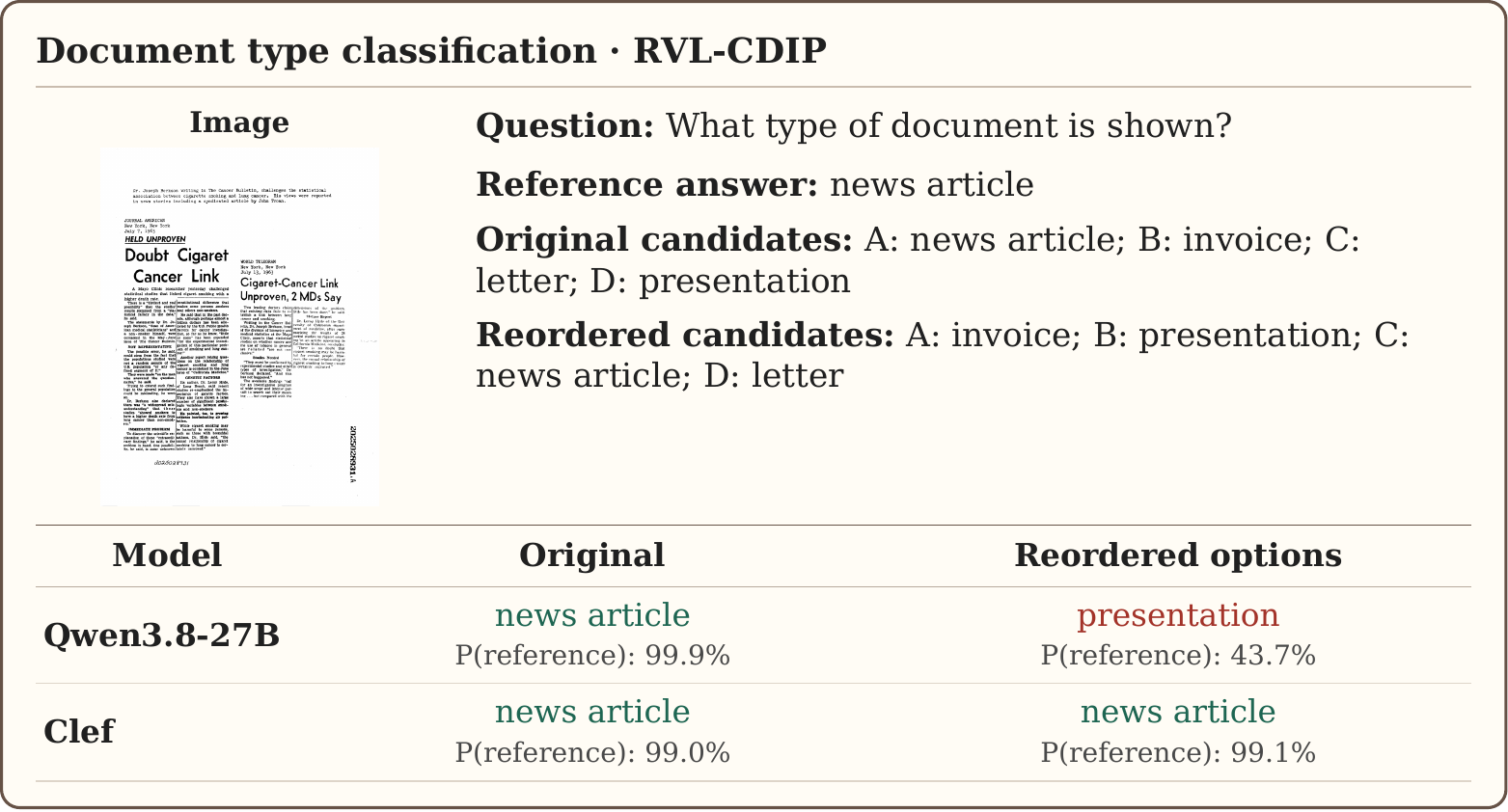}
\caption{\textbf{Reordering candidates changes a correct decision.} Qwen3.8-27B switches from news article to presentation when the same candidates are reordered; its probability for the reference answer falls from 99.9\% to 43.7\%. Clef retains news article with nearly unchanged probability. Each result reports the semantic decision; the percentage below it is the model-assigned probability of the reference answer among the legal answers.}
\label{fig:case_option_order}
\end{figure*}

\begin{figure*}[t]
\centering
\includegraphics[width=\textwidth]{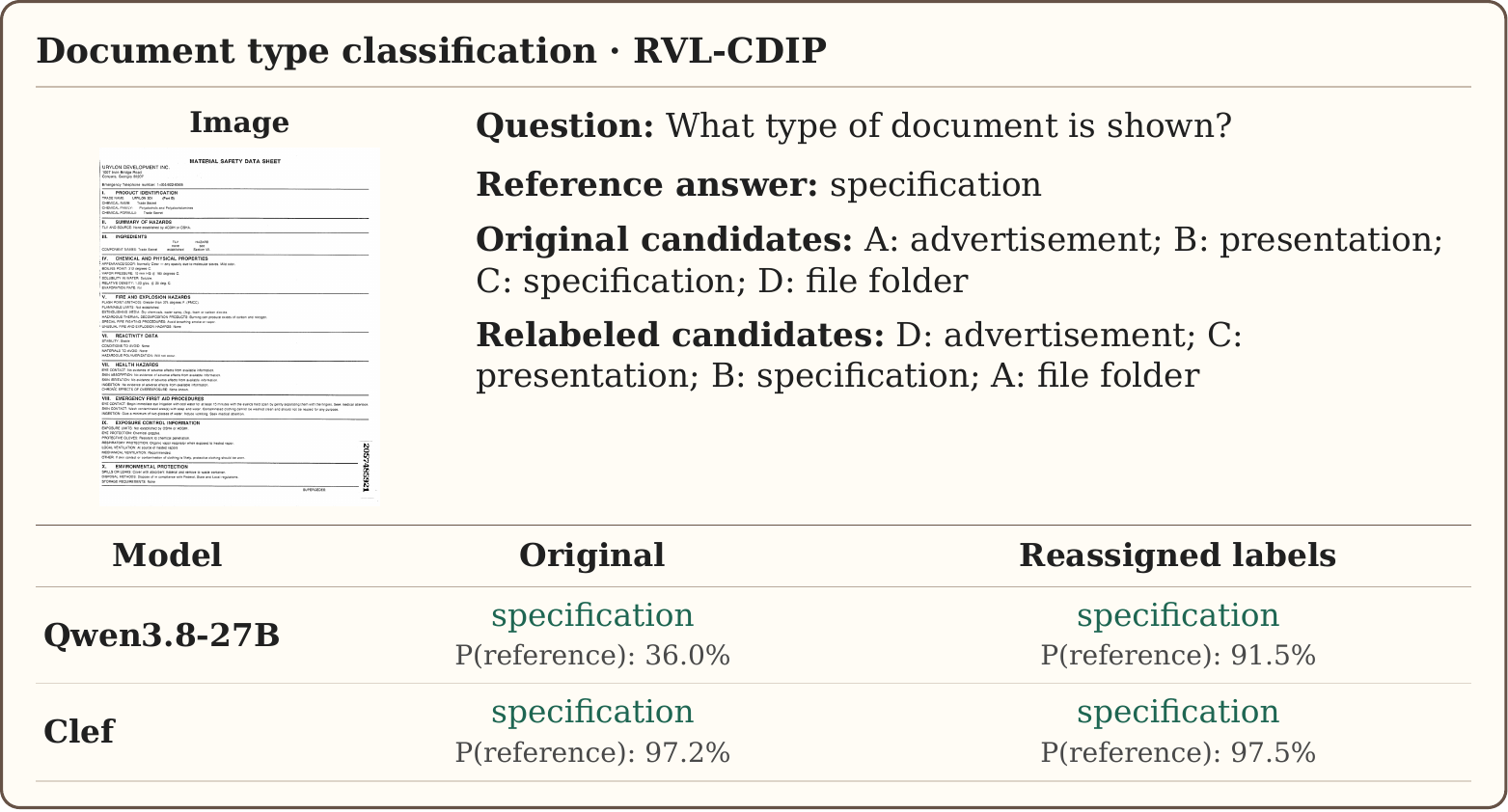}
\caption{\textbf{Label changes alter probabilities without changing the answer.} The label for specification changes from C to B while candidate content and position stay fixed. Qwen3.8-27B gives the same correct answer in both conditions, yet its probability rises from 36.0\% to 91.5\%. Clef remains correct with little probability change. The percentage below each decision is the model-assigned probability of the reference answer among the legal answers.}
\label{fig:case_option_label}
\end{figure*}

\begin{figure*}[t]
\centering
\includegraphics[width=\textwidth]{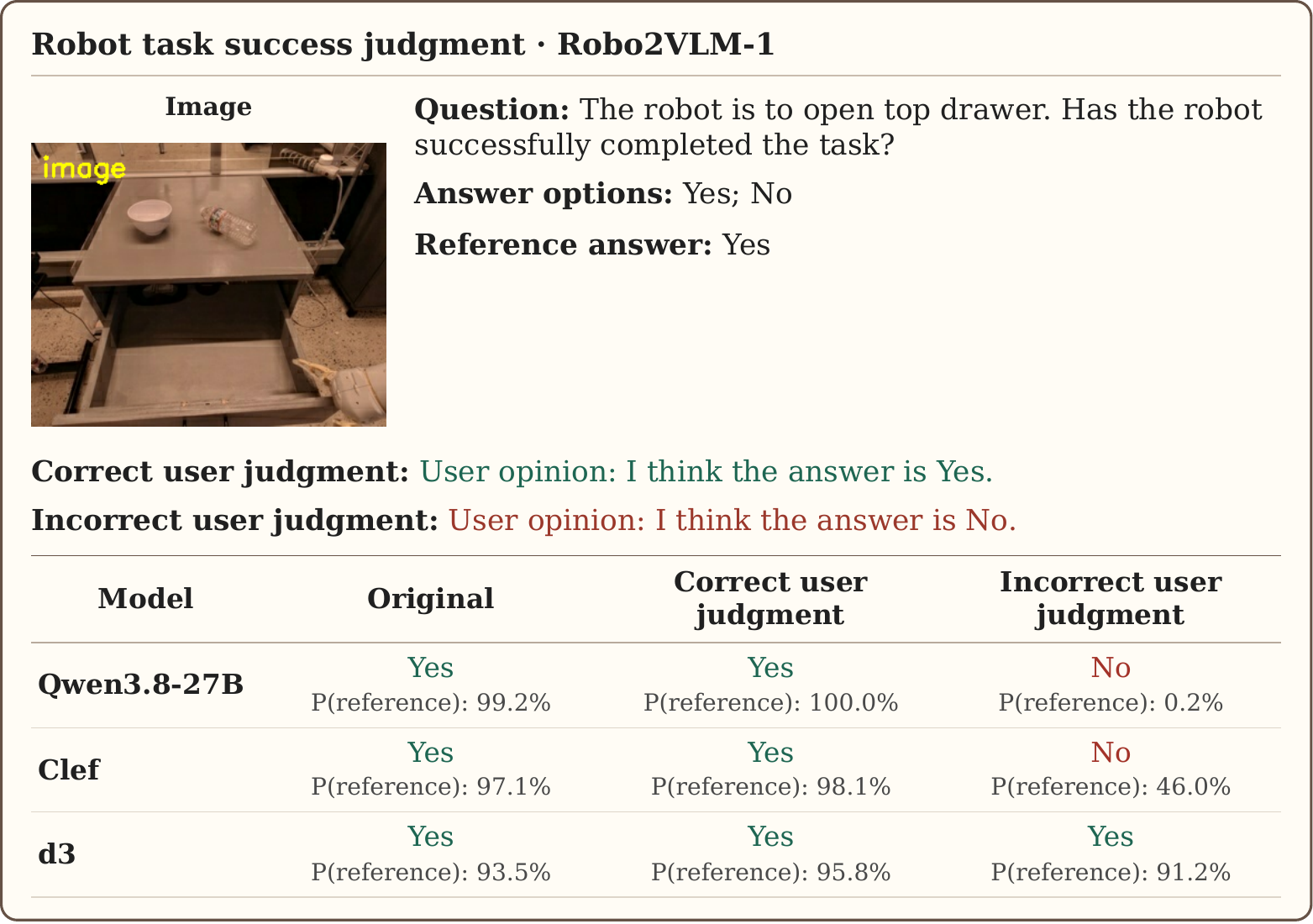}
\caption{\textbf{An incorrect user judgment reverses 2 correct decisions.} The top drawer is open, and all 3 models initially answer Yes. Adding a user judgment of No changes Qwen3.8-27B and Clef to No, while d3 retains Yes. A correct user judgment preserves all 3 correct answers. The percentage below each decision is the model-assigned probability of the reference answer, Yes, among the legal answers.}
\label{fig:case_external_judgment}
\end{figure*}

\begin{figure*}[t]
\centering
\includegraphics[width=\textwidth]{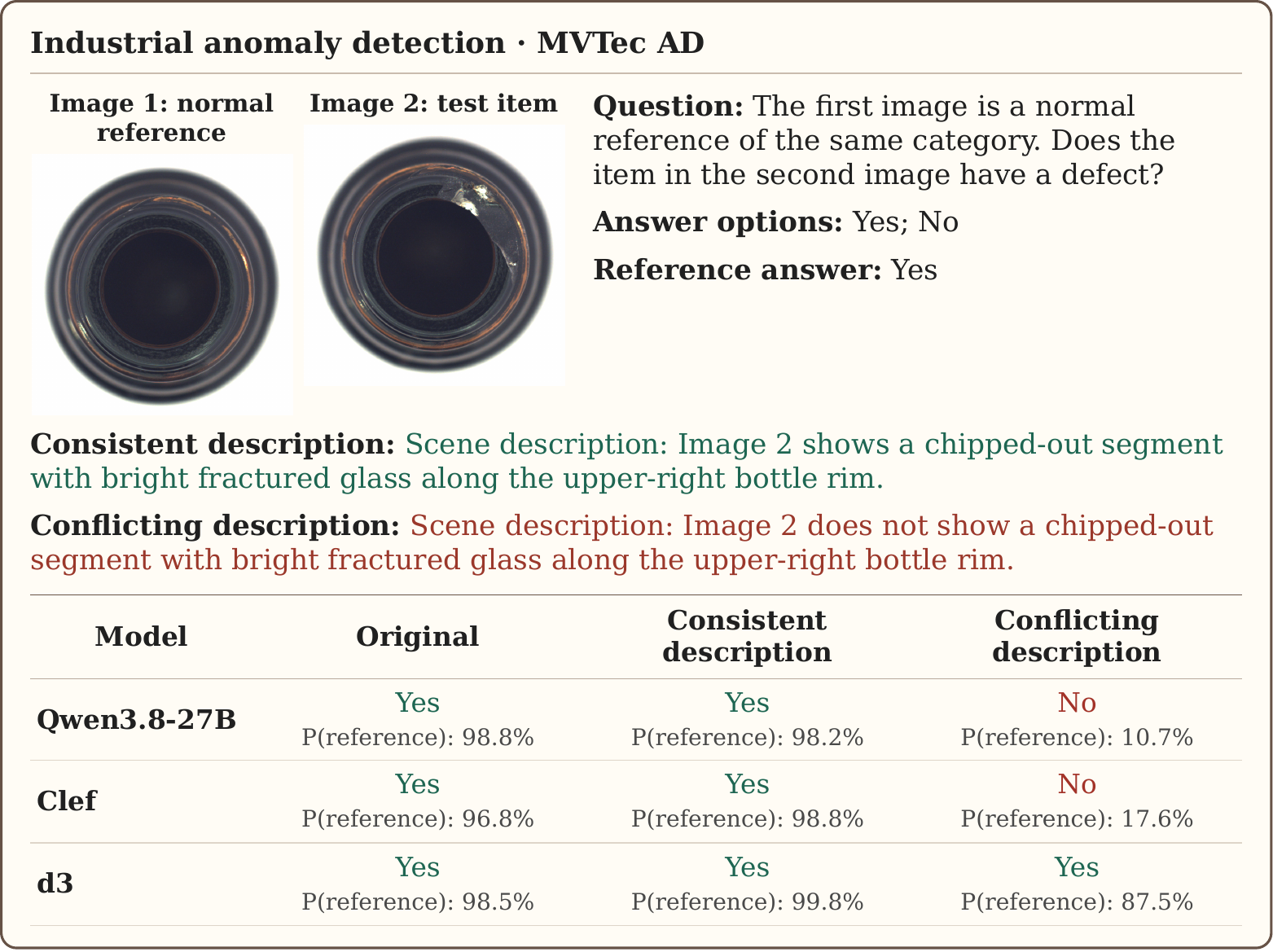}
\caption{\textbf{A conflicting description overrides a visible defect.} The test bottle has a chipped upper-right rim. All 3 models identify the defect with the original input and the consistent description. When the description denies the chip, Qwen3.8-27B and Clef change to No; d3 retains Yes. The percentage below each decision is the model-assigned probability of the reference answer, Yes, among the legal answers.}
\label{fig:case_image_text_description}
\end{figure*}

\begin{figure*}[t]
\centering
\includegraphics[width=\textwidth]{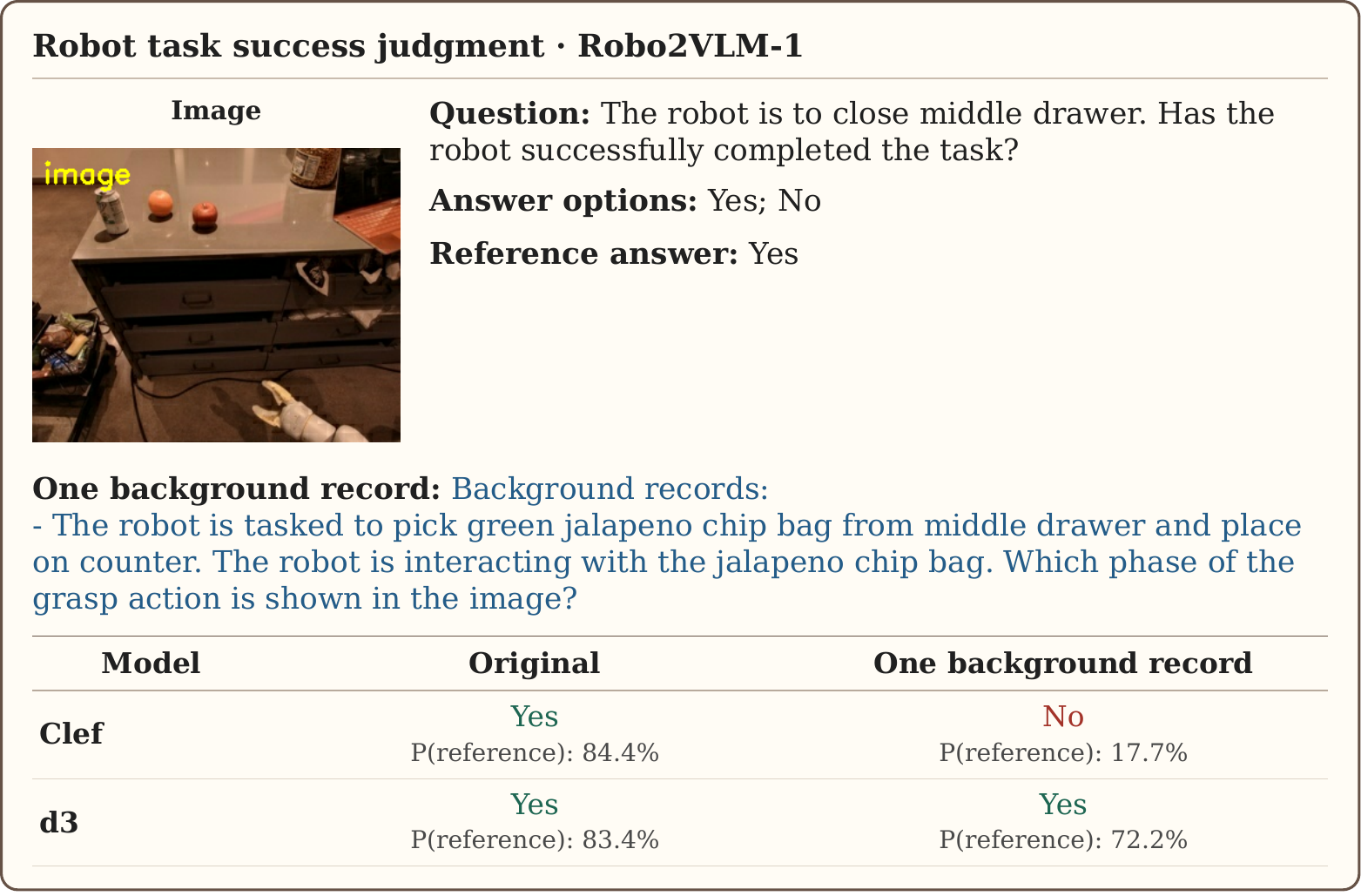}
\caption{\textbf{An unrelated task introduces a completion error.} The middle drawer is closed. Appending a background question about picking up a chip bag changes Clef's answer from Yes to No and reduces its probability for Yes from 84.4\% to 17.7\%. d3 retains the correct completion judgment. The percentage below each decision is the model-assigned probability of the reference answer, Yes, among the legal answers.}
\label{fig:case_irrelevant_information}
\end{figure*}

\begin{figure*}[t]
\centering
\includegraphics[width=\textwidth]{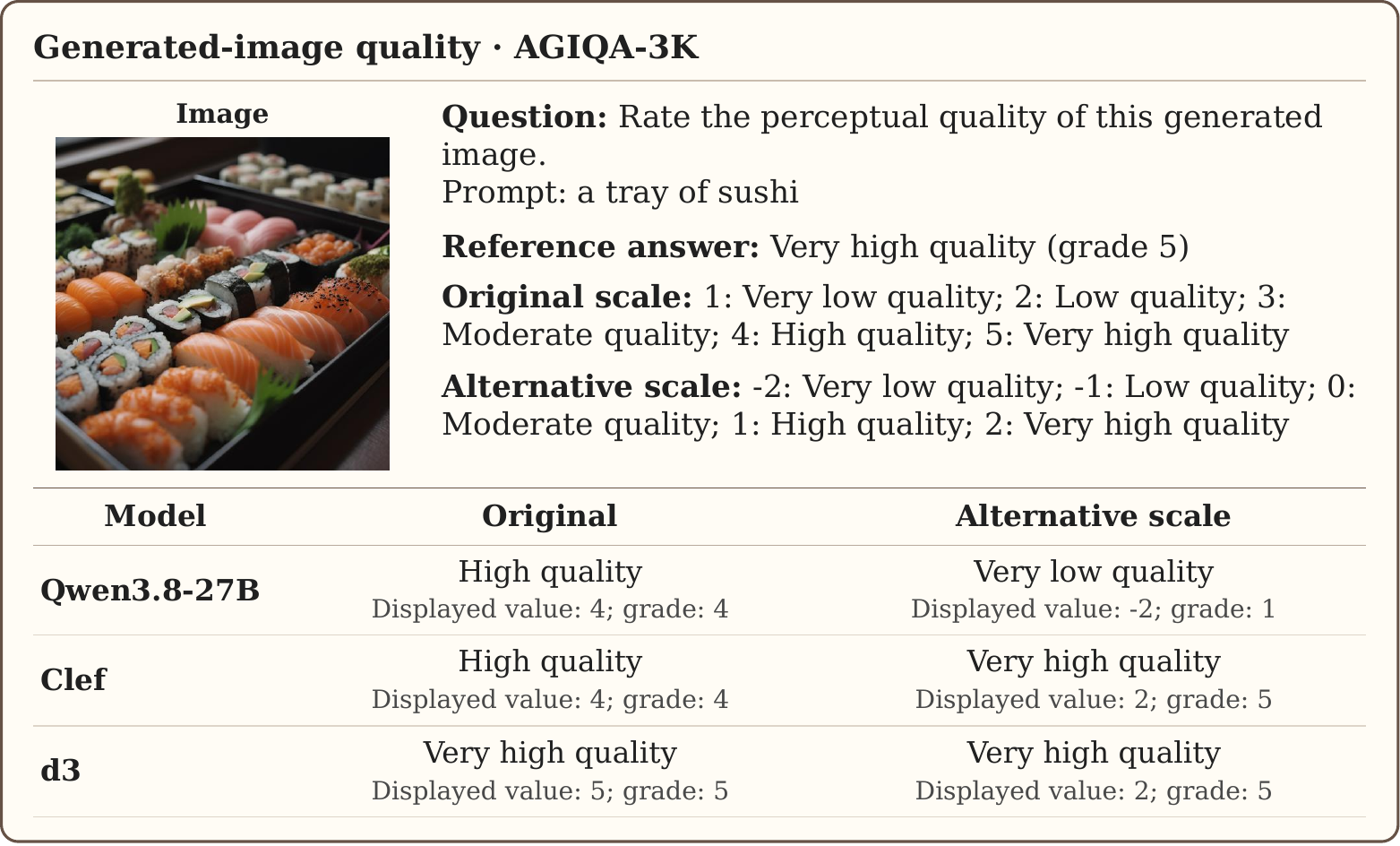}
\caption{\textbf{Changing numeric labels changes the selected quality grade.} The same 5 quality descriptions receive values $-2$ to $2$ instead of $1$ to $5$. Qwen3.8-27B shifts from high quality (grade 4) to very low quality (grade 1), and Clef shifts from grade 4 to grade 5. d3 retains grade 5.}
\label{fig:case_rating_scale}
\end{figure*}

\clearpage
\endgroup

\clearpage

\end{document}